%% file: main.tex
\documentclass[10pt,twocolumn,letterpaper]{article}

\usepackage[accsupp]{axessibility}
\usepackage[pagenumbers]{cvpr} 
\usepackage{multirow}
\usepackage{bbding}
\usepackage{amsmath}
\usepackage{makecell}
\usepackage{algorithm}
\usepackage{algorithmic}
\usepackage{graphicx} 
\usepackage{multicol}
\usepackage{cuted}

\makeatletter
\def\thanks#1{\protected@xdef\@thanks{\@thanks
        \protect\footnotetext{#1}}}
\makeatother

\usepackage{indentfirst}

\definecolor{cvprblue}{rgb}{0.21,0.49,0.74}
\usepackage[pagebackref,breaklinks,colorlinks,allcolors=cvprblue]{hyperref}

\def\paperID{8224} 
\def\confName{CVPR}
\def\confYear{2025}

\title{Pioneering 4-Bit FP Quantization for Diffusion Models: Mixup-Sign Quantization and Timestep-Aware Fine-Tuning}

\author{
Maosen Zhao$^1$$^\ast$, \ 
Pengtao Chen$^1$$^\ast$, \ 
Chong Yu$^2$, \ 
Yan Wen$^1$, \ 
Xudong Tan$^1$, \  
Tao Chen$^1$$^\dagger$
\thanks{$^\dagger$Corresponding author.\quad$^\ast$Equal contribution.} \\
$^1$ School of Information Science and Technology, Fudan University \\
$^2$ Academy for Engineering and Technology, Fudan University \\
{\tt\small 20307130202@fudan.edu.cn, eetchen@fudan.edu.cn}
}

\begin{document}
\maketitle
\input{sec/0_abstract}    
\input{sec/1_intro}

\input{sec/2_related_work}
\input{sec/3_observation}

\input{sec/4_methodology}
\input{sec/5_experiment}
\input{sec/6_conclusion}

\input{sec/7_acknowledge}
{
    \small
    \bibliographystyle{ieeenat_fullname}
    \bibliography{main}
}
\input{sec/X_suppl}
\end{document}

%% file: sec/0_abstract.tex
\begin{abstract}

Model quantization reduces the bit-width of weights and activations, improving memory efficiency and inference speed in diffusion models. However, achieving 4-bit quantization remains challenging. Existing methods, primarily based on integer quantization and post-training quantization fine-tuning, struggle with inconsistent performance. Inspired by the success of floating-point (FP) quantization in large language models, we explore low-bit FP quantization for diffusion models and identify key challenges: the failure of signed FP quantization to handle asymmetric activation distributions, the insufficient consideration of temporal complexity in the denoising process during fine-tuning, and the misalignment between fine-tuning loss and quantization error. To address these challenges, we propose the mixup-sign floating-point quantization (MSFP) framework, first introducing unsigned FP quantization in model quantization, along with timestep-aware LoRA (TALoRA) and denoising-factor loss alignment (DFA), which ensure precise and stable fine-tuning. Extensive experiments show that we are the first to achieve superior performance in 4-bit FP quantization for diffusion models, outperforming existing PTQ fine-tuning methods in 4-bit INT quantization.

\end{abstract}

%% file: sec/1_intro.tex
\section{Introduction}
\label{sec:intro}



Despite the impressive performance of diffusion models (DMs) in image generation~\cite{yu2015lsun, rombach2022high}, their computational and memory demands, particularly for high-resolution outputs, pose significant challenges for deployment on resource-constrained edge devices, highlighting the need for model compression to address these limitations. Model quantization, a key technique within model compression, reduces the bit-width of model weights and activations, typically stored in 32-bit format, to lower precision. This reduction lowers memory usage and accelerates inference speed. By decreasing the bit-width, quantization improves both temporal and memory efficiency in mainstream models, maintaining robust performance, particularly in resource-constrained environments ~\cite{liang2021pruning,li2024contemporary,huang2024empirical}.

The current quantization methods for diffusion models can be broadly categorized into two primary approaches. The first, post-training quantization (PTQ), optimizes the quantization parameters after the model has been trained, typically by minimizing the quantization error \cite{shang2023post, li2023q}. While PTQ is effective for 4-bit quantization of weights, it is limited by its reliance on 8-bit quantization for activations, as further reduction in activation bit-width leads to significant performance degradation. In contrast, quantization-aware training (QAT) integrates quantization into the training process, enabling the model to learn with 4-bit precision from scratch \cite{li2024q}. Although QAT can achieve high-performance 4-bit models, it incurs substantial computational overhead, rendering it less practical for many real-world applications~\cite{krishnamoorthi2018quantizing,esserlearned}.

To achieve fully quantized 4-bit diffusion models with minimal overhead, fine-tuning has emerged as a promising solution. This approach leverages pre-trained models and adjusts a small subset of parameters to narrow the performance gap between the quantized model and its full-precision counterpart. While some studies have explored fine-tuning for 4-bit quantization in diffusion models \cite{he2023efficientdm, wang2024quest}, these methods often fail to achieve consistent performance under standard configurations (e.g., quantizing all layers but not some). Consequently, developing a universally effective and scalable fine-tuning method for 4-bit quantization in diffusion models remains an open challenge.

To be noticed, existing quantization methods for diffusion models primarily rely on integer (INT) quantization, which has long been the dominant approach. However, recent developments have demonstrated the considerable potential of floating-point (FP) quantization. Compared to INT quantization, FP quantization offers greater flexibility in modeling complex weight and activation distributions~\cite{micikevicius2022fp8}, leading to improved performance in visual tasks at 8-bit precision ~\cite{van2023fp8, kuzmin2022fp8}, and remarkable results in large language models under 4-bit precision \cite{liu2023llm,wang2024fp4}. Moreover, FP quantization provides significant advantages in inference acceleration, with NVIDIA H100 achieving a 1.45× speedup with FP8 quantization, outperforming INT8 \cite{nvidia2024blackwell,zhang2024integer}. Despite these advantages, the application of low-bit FP quantization in diffusion models remains largely unexplored, presenting a promising avenue for future research.

In summary, to address the challenges in achieving 4-bit diffusion models, we propose a baseline method using a search-based signed FP quantization framework~\cite{liu2023llm, chen2024low} combined with single-LoRA fine-tuning, aiming to realize 4-bit FP quantized diffusion models. Through this exploration, we uncover several findings that are significant for both diffusion model quantization and FP quantization:
(1) There exists many layers which exhibit asymmetric distributions, due to the nonlinear activation function $SiLU$. The application of traditional signed FP quantization with a symmetric distribution leads to significant precision loss in the sub-zero area, causing substantial performance degradation after quantization. (2) Fine-tuning is conducted based on the denoising process, which is regarded as a complex task involving the restoration from outlines to details~\cite{wang2023diffusion}. However, current methods typically apply a single LoRA to fine-tune severely degraded models across all timesteps, which leads to suboptimal learning at certain timesteps. (3) Predicted noise plays a varying role at different timesteps during denoising, which is the key to diffusion models. Arising from the neglect of this variation across timesteps, we identify a mismatch between the impact of quantization and the loss function in current methods, which undermines the effectiveness
of fine-tuning.

Facing the challenges in achieving 4-bit FP quantization, we propose constructive strategies: (1) To handle different distribution effectively, we propose a mixup-sign floating-point quantization (MSFP) framework, where unsigned FP quantization with an added zero point leads to a more compatible distribution of discrete points with the anomalous distributions in activations and signed FP quantization continues to exhibit strong representation capacity on other distributions. (2) Realizing that the current single-LoRA fine-tuning approach is in lack of flexibility across timesteps, we introduce timestep-aware LoRA (TALoRA), incorporating multiple LoRAs and a timestep-aware router to dynamically select the appropriate LoRA for each timestep in the denoising process. (3) To further improve the effectiveness of fine-tuning, we introduce a denoising-factor loss alignment (DFA), ensuring the loss function, and the guidance of the fine-tuning, are consistent with the actual quantization deterioration across timesteps. 

In summary, our contributions are as follows:
\begin{enumerate}[label=(\roman*)]
    \item 
    We are the first to identify that signed FP quantization struggles with asymmetric activations, which arise from the nonlinear behavior of activation functions. To address this, we introduce the MSFP framework, which is also the first effective application of unsigned FP quantization in quantization, offering a novel approach for achieving low-bit quantization.
    \item 
    For the fine-tuning of low-bit diffusion models, which is based on the denoising process, we precisely define it as a multi-task process across timesteps and introduce an efficient TALoRA module. Furthermore, we improve the alignment of loss function with quantization error via DFA strategy, enabling stable and reliable fine-tuning to achieve low-bit diffusion models.
    \item 
    We focus on identifying and eliminating three major barriers to effective low-bit FP quantization in diffusion models. Extensive experiments on DDIM and LDM demonstrate that our work achieves SOTA results for 4-bit quantization. The proposed MSFP, TALoRA and DFA have greatly advanced the progress of low-bit quantization in diffusion models.
    
\end{enumerate}

%% file: sec/2_related_work.tex
\section{Related Work}
\subsection{Diffusion Model Quantization}

\begin{figure*}[htbp]
    \centering
    \includegraphics[width=\textwidth]{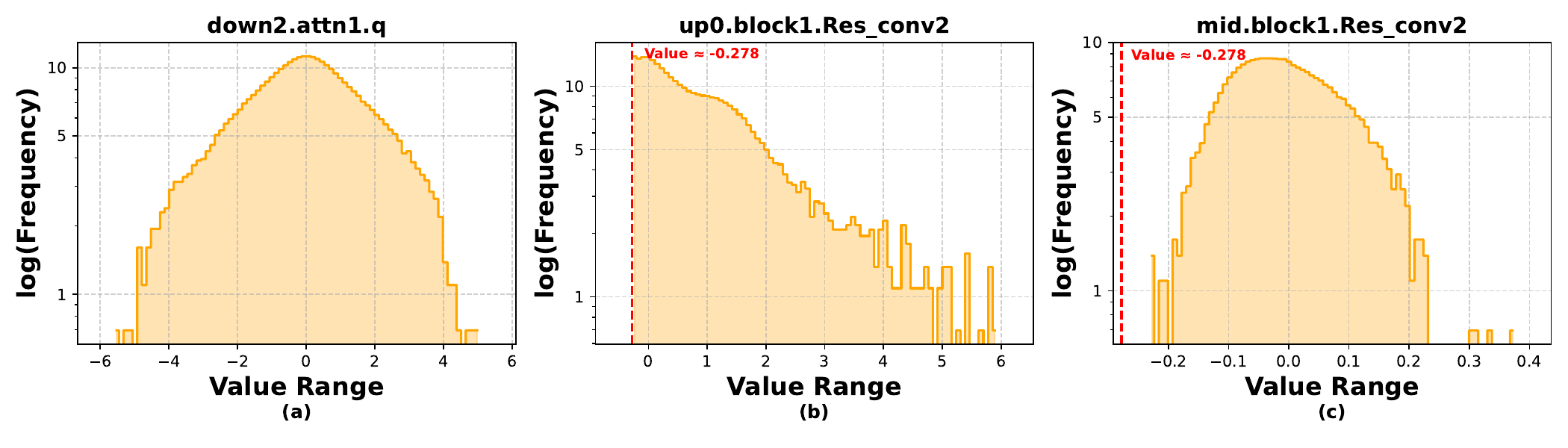}
    \vspace{-25pt}
    \caption{The activation distributions in NALs and AALs, results on the CelebA dataset.
(a) The paradigm of NALs with symmetric activations.
(b) The typical paradigm of AALs with asymmetric activations, where unsigned FP quantization is more suitable.
(c) The infrequent paradigm of AALs with relatively symmetric activations, where either signed or unsigned FP quantization could be applicable.
}
    \label{fig:3_activations}
\end{figure*}

There are two main approaches in the quantization of diffusion models: QAT~\cite{esserlearned,jacob2018quantization} and PTQ~\cite{nagel2020up}. QAT is particularly effective for low-bit quantization while it re-trains the model from scratch, consuming extensive computational resources and time~\cite{li2024q}. In contrast, PTQ offers greater time efficiency, making it more practical for large models in real-world applications.
Recent advancements in PTQ have concentrated on optimizing calibration datasets to enhance reconstruction accuracy~\cite{shang2023post, li2023q, he2024ptqd, liu2024enhanced} and mitigating quantization errors stemming from the temporal and structural properties of diffusion models~\cite{li2023q, chu2024qncd,so2024temporal,sun2024tmpq}. To further achieve fully 4-bit quantization, fine-tuning techniques has been introduced in PTQ-based quantization. EfficientDM~\cite{he2023efficientdm} develops a LoRA-based fine-tuning framework, while QuEST~\cite{wang2024quest} focuses on optimizing quantization-unfriendly activations. Both works stagnate in addressing data distribution during fine-tuning, failing to account for the specific challenges involved in fine-tuning within the context of diffusion model quantization.

Meanwhile, recent advancements in FP quantization have made significant strides in model quantization~\cite{liu2023llm, kuzmin2022fp8}, highlighting its potential for diffusion models. While existing low-bit quantization methods perform well for linear models, applying FP quantization to convolutional diffusion models remains more challenging. To date, only one study has explored 8-bit activation quantization in diffusion models using a basic search-based approach~\cite{chen2024low}, with no research addressing lower-bit quantization. This underscores both the challenges and the untapped potential of achieving 4-bit FP quantization for diffusion models.

\subsection{Parameter-Efficient Fine-Tuning}
Parameter-efficient fine-tuning (PEFT) has emerged as an effective alternative to full model fine-tuning, focusing on adjusting only a subset of parameters while keeping the majority frozen, thereby reducing storage overhead. Low-rank adapters (LoRA)~\cite{hu2021lora}, originally developed for large language models, have become one of the most widely used PEFT methods. Leveraging LoRA's strong transferability, QLoRA~\cite{dettmers2023qlora} can be effectively applied to fine-tune low-bit diffusion models. However, previous LoRA-based fine-tuning is suboptimal in practice. In this paper, we make a further exploration on this and incorporate timestep-level adaptation inspired by MoELoRA~\cite{feng2024mixture}, enhancing performance of low-bit quantized DMs.


%% file: sec/3_observation.tex
\section{Challenges \& Exploration}
\subsection{Preliminary}
\textbf{Diffusion Models.}
The diffusion model is a new generation framework that completes learning by adding noise and completes generation in a denoising manner.
As for the forward process, the noise is injected into ground-truth images $\boldsymbol{x}_0$ at random timesteps. This enables the DMs to learn the distributions of noise through noisy images $\boldsymbol{x}_{t}$, which can be obtained as follows~\cite{ho2020denoising}:

\begin{equation}
    \boldsymbol{x}_{t}=\sqrt{\overline{\alpha }_{t}}\boldsymbol{x}_{0}+\sqrt{1-\overline{\alpha }_{t}}\boldsymbol{\varepsilon}.
\end{equation}

Here $\boldsymbol{\varepsilon}$ represents a standard Gaussian noise and $\overline{\alpha }_{t}$ is the accumulated noise intensity, calculated by:

\begin{equation}
    \overline{\alpha} _{t}=\prod ^{t}_{i=1}\alpha _{i},
\end{equation}

\noindent where $\alpha_{s}$ governs the noise intensity under each timestep. Once the DMs are well-trained, the Gaussian noise image will be inputted to the DMs, and undergo the iterative denoising process. Specifically, the noise can be predicted by DMs and used to obtain the image $\boldsymbol{x}_{t}$ under timestep $t$, which could range from $T$ to $0$, with the objective of obtaining the image $\boldsymbol{x}_{t-1}$ in a superior quality:

\begin{equation}
    \boldsymbol{x}_{t-1}=\dfrac{1}{\sqrt{\alpha _{t}}}\left( \boldsymbol{x}_{t}-\dfrac{1-\alpha _{t}}{\sqrt{1-\overline{\alpha }_{t}}}\cdot \boldsymbol{\varepsilon} _{\theta }\left( \boldsymbol{x}_{t},t\right) \right) +\sigma _{t}\boldsymbol{\delta},
    \label{equation3}
\end{equation}

\noindent where $\boldsymbol{\varepsilon} _{\theta }\left( \boldsymbol{x}_{t},t\right)$ is the predicted noise at timestep $t$ and $\boldsymbol{\delta}$ is a newly added noise with the factor $\sigma _{t}$ to ensure diverse results. Here we define a denoising factor $\gamma _{t}$, which is formulated as:

\begin{equation}
    \gamma _{t}=\dfrac{1}{\sqrt{\alpha _{t}}}\cdot \dfrac{1-\alpha _{t}}{\sqrt{1-\overline{\alpha }_{t}}}.
\end{equation}

According to Equation~\ref{equation3}, $\gamma _{t}$ indicates the impact of the prediction noise under timestep $t$. The greater the factor $\gamma _{t}$, the stronger the predicted noise effect in the denoising. 

\textbf{Model Quantization.}
The fundamental paradigm of model quantization is the process of transforming a continuous data distribution into a finite set of discrete points. Consequently, the quality of the quantized model is inextricably related to the distribution of discrete points. According to the discrete point type, quantization is defined as two categories, widely-used INT quantization and emerging FP quantization.
Analogous to the process of INT quantization, a floating-point vector $\boldsymbol{x}$ can be quantized as follows:

\begin{equation}
    \widehat{\boldsymbol{x}}=Clip\begin{pmatrix}
\lfloor \dfrac{\boldsymbol{x}}{s}\rceil +z, l, u
\end{pmatrix}\cdot s.
\label{5}
\end{equation}

\begin{figure*}[ht]
    \begin{minipage}[t]{0.33\linewidth}
        \centering
        \includegraphics[width=1\linewidth]{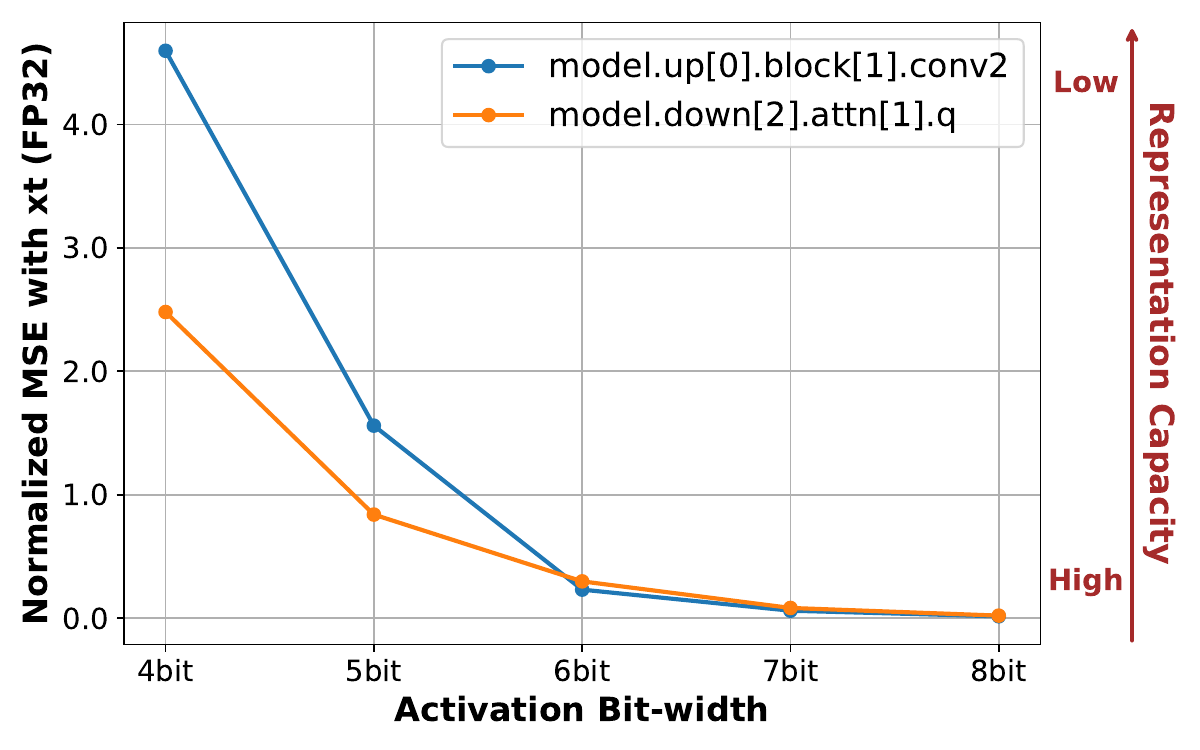}
        \captionsetup{justification=centering} 
        \caption{
    Effect of bit-width reduction on activation representation capacity in AALs (blue) and NALs (orange) under signed FP quantization, evaluated on CelebA dataset.}

    \label{fig:sign}
    \end{minipage}%
    \begin{minipage}[t]{0.33\linewidth}
        \centering
        \includegraphics[width=1\linewidth]{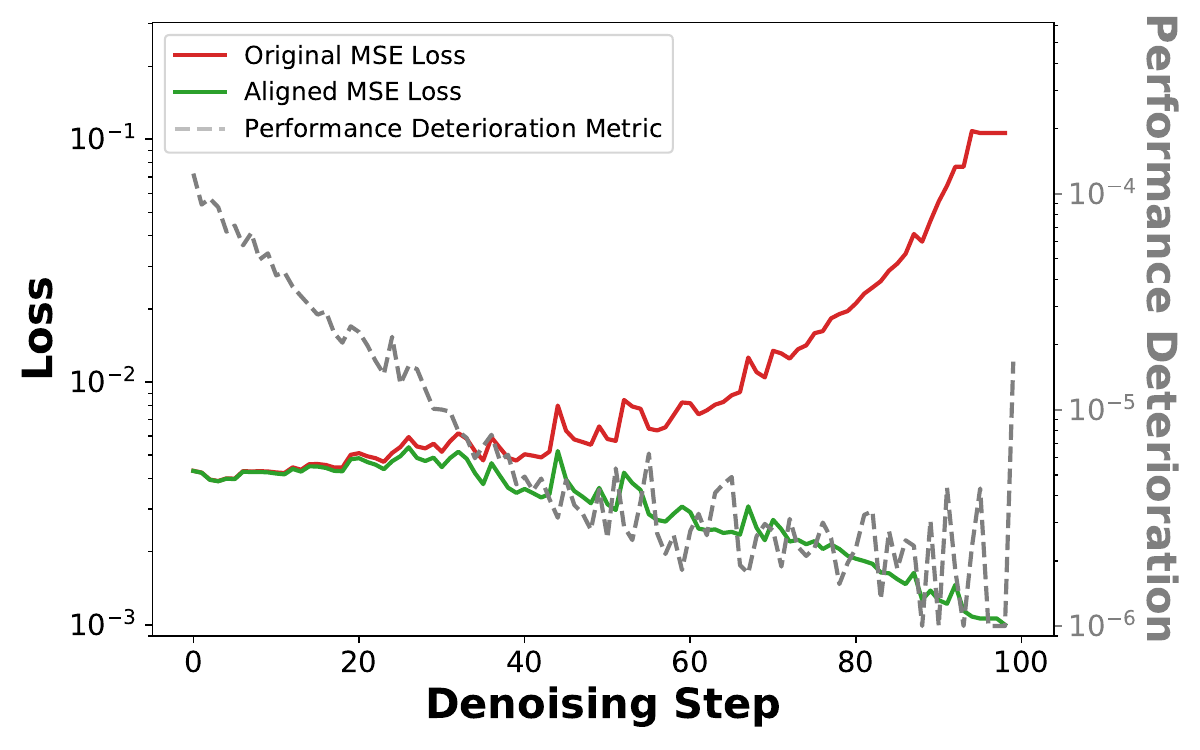}
        \caption{Two loss, and performance degradation between the quantized and full-precision models across steps. Compared with metric, the original loss shows an inverse trend, while the aligned loss remains consistent.}
    \captionsetup{justification=centering} 
        \label{fig:3_loss}
    \end{minipage}
    \begin{minipage}[t]{0.05\linewidth}
    \end{minipage}
    \begin{minipage}[t]{0.33\linewidth}
        \centering
        \includegraphics[width=1\linewidth]{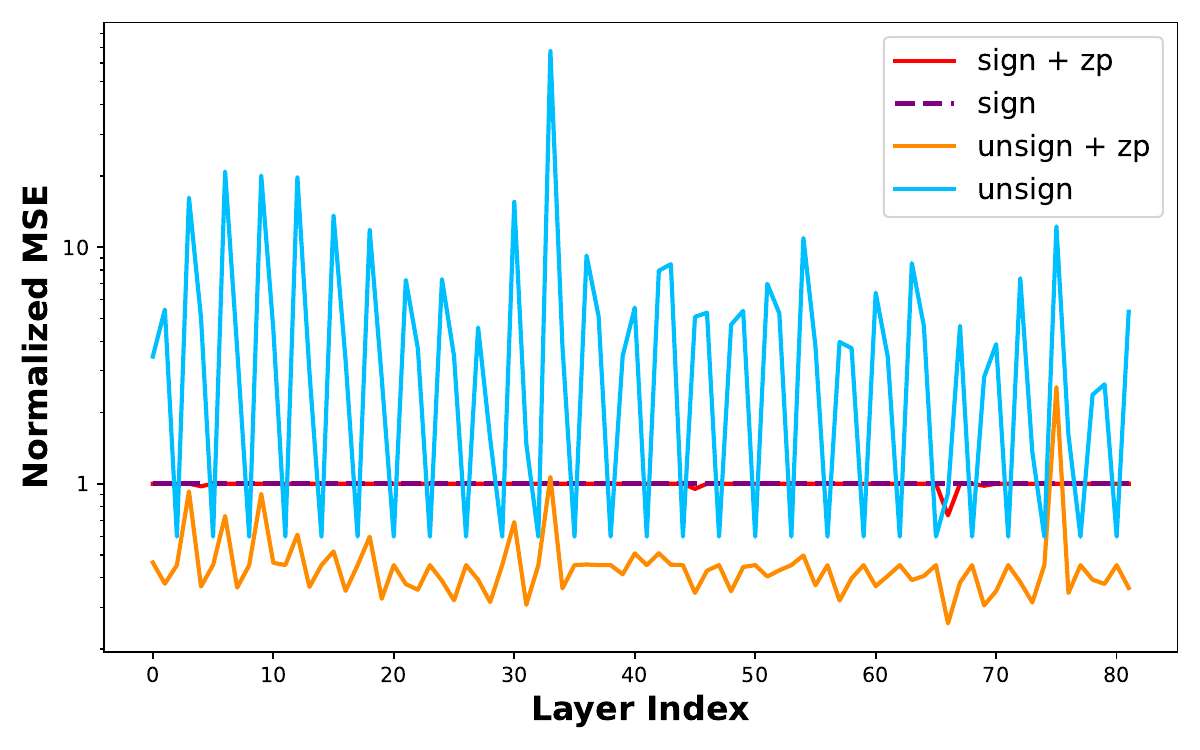}
        \caption{The MSE of activations before and after quantization across all AALs under four different strategies, normalized against the baseline of signed FP quantization without zero point (purple).}
        \label{fig:unsign_mse}
    \end{minipage}
\end{figure*}

Here $\lfloor \rceil$ is the rounding operation. $l$ and $u$ are the minimum and maximum quantization thresholds while scaling factor $s$ and zero-point $z$ together constitute the quantization parameters.
As shown in Equation \ref{5}, INT quantization results in an evenly spaced distribution of discrete points. While INT quantization is straightforward to implement, it may be too simplistic for continuous distributions that have significant variations in density, where evenly spaced intervals in INT quantization are not effective.
On the contrary, the discrete points under FP quantization are not uniformly spaced in distribution, as shown below~\cite{liu2021improving}:

\begin{equation}
    f=\left( -1\right) ^{s}2^{p-b}\left( 1+\dfrac{d_{1}}{2}+\dfrac{d_{2}}{2^{2}}+\ldots \dfrac{d_{m}}{2^{m}}\right) ,
    \label{equation6}
\end{equation}

\noindent where $s$ is the sign bit, $d_{i}$ is the $m$-bit mantissa, $p$ is the $e$-bit exponent and $b$ is the bias that serves as both scaling factor and threshold in INT quantization. Previous studies in FP quantization rely on signed FP quantization, where $s$ is set to 1. For an $n$-bit FP quantization, the bit-width is distributed across the mantissa, exponent, and sign bit, with the condition
    $m+e+s=n$.
The different combinations of $m$ and $e$ allow FP quantization to represent data in multiple formats with a fixed number of bits, denoted as $E_iM_j$, where $i$-bit exponent and $j$-bit mantissa are specified. A larger $j$ results in higher precision within each interval, while a larger $i$ expands the range of covered intervals.

There appears to be an inherent suitability of FP quantization for DMs, as the majority of weights and activations follow a normal distribution symmetric around $value=0$. This aligns well with the unevenly spaced distribution, where discrete points are dense in the small-value region and sparse in the large-value region, in FP quantization~\cite{zhang2024integer}. Additionally, the flexible formatting mechanism allows FP quantization to better accommodate complex distribution scenarios.

\subsection{Barriers to Effective Low-Bit FP Quantization in Diffusion Models}
\label{section3.2}
In order to implement high-performing 4-bit diffusion models, we implemented FP quantization that is more compatible with the data distribution of diffusion models, with the search-based strategy in previous work \cite{liu2023llm}. Building on this, we deployed a LoRA-based fine-tuning strategy to address the performance degradation caused by 4-bit quantization. Despite this, we find that the performance of the 4-bit FP diffusion model remains satisfactory and we identify two main issues that we are facing: 
(1) For FP quantization, although FP quantization works well at 8 bits, it experiences a sharp performance degradation at 4 bits. \textit{How to improve the ability of FP quantization to represent data under low-bit quantization}?
(2) For standard post-training LoRA-based fine-tuning, exhibits instability and suboptimal results when applied to low-bit quantized diffusion models. \textit{How can we make LoRA more efficient and accurate in learning the loss information at different denoising timesteps}? In the following parts, we will explore the underlying causes of these two issues and the feasible strategies to address them.

\textbf{Observation 1: Previous signed FP quantization fails to achieve effective low-bit quantization in Activation-Anomalous Layers.}

In diffusion models, we observed that the nonlinear activation layer $SiLU$, defined as $SiLU\left( \boldsymbol{x}\right) =\dfrac{\boldsymbol{x}}{1+e^{-\boldsymbol{x}}}$, 
is commonly situated between layers.

$SiLU$ causes the abnormal activations for the subsequent layer. As depicted in Panel (b) of Figure \ref{fig:3_activations}, all values below 0 are compressed into the range of $[-0.278, 0)$. In this paper, we refer to layers with such asymmetric activations as Anomalous-Activation-Distribution Layers (AALs) and the other layers as Normal-Activation-Distribution Layers (NALs).
Mainstream signed FP quantization typically sets the maximum threshold of the quantizer high to accurately represent normal positive activations. However, this approach results in a significant precision loss when dealing with values below 0. 
Figure \ref{fig:sign} illustrates the representation capacity of FP signed quantization in both NALs and AALs under different bit widths. When the bit width drops below 6 bits, AALs suffer more severe performance degradation compared to NALs, which ultimately leads to the failure in low-bit FP quantization. This phenomenon suggests that mitigating the performance decline in AALs is a critical step towards improving low-bit FP quantization in diffusion models.



\textbf{Observation 2: The single-LoRA-based strategy is overly simplistic for fine-tuning quantized diffusion models across different timesteps.}


Previous work has focused on adapting LoRA for the quantization of diffusion models but has not fully explored LoRA's performance in the context of denoising. Considering that the denoising process of diffusion models starts with recovering outlines and progresses to restoring details, we question whether the single-LoRA fine-tuning strategy can handle this complexity. In Table \ref{tab:2LORA}, we compare the baseline model, which uses a single-LoRA strategy for fine-tuning, with two alternative strategies. The second strategy assigns a separate LoRA for the first and last 50 timesteps, resulting in a significant improvement over the baseline.  In contrast, the third strategy also introduces a dual-LoRA strategy, it randomly selects one for each timestep, resulting in much worse performance.
These results suggest that applying multiple LoRAs, allocated in a structured manner across timesteps, enhances model performance, while disordered selection of LoRAs could lead to suboptimal results.
This motivates us to approach the fine-tuning of quantized diffusion models as a multi-task process and assign multiple LoRAs across different timesteps with a rational approach in allocation.

\begin{figure*}[t]
    \centering
    \includegraphics[width=0.95\linewidth]{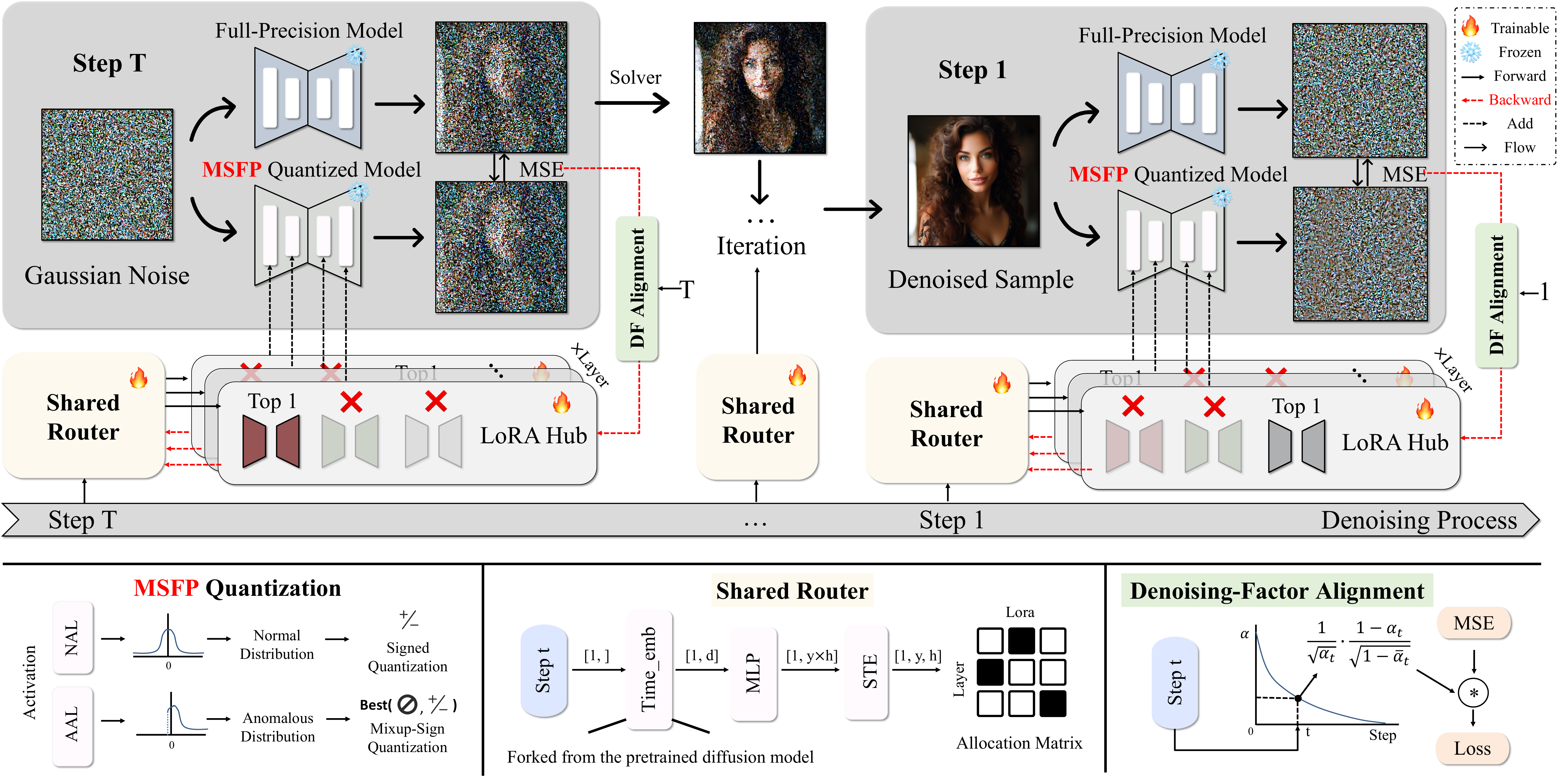}
    \caption{The pipeline of our proposed method. UNets are applied to the Mixup-Sign Floating-Point Quantization (MSFP), where distinct floating-point quantization schemes are employed for Anomalous-Activation-Distribution Layers (AALs) and Normal-Activation-Distribution Layers (NALs). During the fine-tuning stage, multiple LoRA modules are introduced, and a timestep-aware routing mechanism is used for dynamic LoRA allocation across different timesteps. Additionally, a denoising-factor alignment technique is employed to align the loss function with quantization-induced performance degradation.}
    \label{fig:pipeline}
\end{figure*}

\textbf{Observation 3: The MSE of the predicted noise of full-precision and quantized models does not reflect the actual impact of quantization at different timesteps.}

In fine-tuning, a commonly used loss function calculates the MSE between the noise predictions of the full-precision and quantized diffusion models, using denoised images from the full-precision model at the previous timestep as inputs:

\begin{equation}
    L_{\boldsymbol{\varepsilon} _{\theta }}^{t}= \left\| \boldsymbol{\varepsilon} _{\theta}\left( \boldsymbol{x}_{t},t\right) -\widehat{\boldsymbol{\varepsilon} }_{\theta}\left( \boldsymbol{x}_{t},t\right) \right\| ^{2}.
\end{equation}

By observing the variation in this loss during single-LoRA fine-tuning (see Figure \ref{fig:3_loss}), we identify an unexpected trend: the loss increases progressively faster as denoising advances. This contradicts the principle of denoising, where image quality should improve with each step, and the impact of predicted noise should diminish over time. By the final step, the impact of quantization on model performance should be negligible, as the predicted noise no longer affects the input. However, the loss is at its maximum, indicating that quantization error is most significant at this stage, contradicting the expectation that its influence should diminish over time.

To highlight the discrepancy between the loss and actual quantization errors, we define the performance gap at each step as the difference in denoising quality between the quantized and full-precision models, as the ultimate goal of denoising is to yield high-quality images.
We input the previous output image $\boldsymbol{x}_t$ from the full-precision model and calculate the performance gap between the denoised image $\boldsymbol{x}_{t-1}$ from the full-precision model and the denoised image $\widehat{\boldsymbol{x}}_{t-1}$ from the quantized model, measured by
$MSE(\boldsymbol{x}_{t-1},\widehat{\boldsymbol{x}}_{t-1})$.
As shown in Figure \ref{fig:3_loss}, this misalignment between the loss and the actual performance gap leads to deviations in LoRA’s learning. This necessitates aligning our loss function with the denoising process during fine-tuning.

%% file: sec/4_methodology.tex
\section{Methodology}
In this section, we explore the issues identified in Section~\ref{section3.2} and propose corresponding solutions. As illustrated in Figure~\ref{fig:pipeline}, we introduce a mixup-sign FP quantization framework to address the diverse activation distributions in the first stage. During fine-tuning, the timestep-aware routing mechanism and denoising-factor loss alignment work in tandem to enable high-quality learning, ultimately enabling the realization of optimized 4-bit FP diffusion models.

\begin{table}
    \centering
    \setlength{\tabcolsep}{0.5 mm}{
    \begin{tabular}{ccc}
    \toprule
         \textbf{Method} & \textbf{Bits (W/A)} & \textbf{FID $\downarrow$} \\
    \midrule
         FP & 32/32 & 6.49 \\
         Single-LoRA & 4/4 & 19.41 \\
         Dual-LoRA (Split Steps in Half) &4/4 & 17.07 \\
         Dual-LoRA (Random Allocation) &4/4 & 41.96 \\
    \bottomrule     
    \end{tabular}}
        \caption{The impact of the number of LoRAs and their allocation across timesteps on the performance of the fine-tuning. The results is evaluated by 4-bit quantization on CelebA dataset.}
    \label{tab:2LORA}
\end{table}

\subsection{Mixup-Sign Floating Point Quantization}
To address the challenges of low-bit FP quantization failure in AALs, we leverage FP quantization by allocating more discrete points to areas with high data concentration. Motivated by the half-normal distribution of activations in AALs, we introduce unsigned floating-point quantization. However, as shown in Equation \ref{equation6}, when using unsigned FP quantization with $s$ set to 0, we round all data in the sub-zero range to zero, losing important negative information. To address this, we introduce a zero point in the range of [-0.278, 0) to recover most sub-zero activations. The updated quantization formula becomes:

\begin{equation}
    f_{unsign}=\left( -1\right) ^{s}2^{p-b}\left( 1+\dfrac{d_{1}}{2}+\dfrac{d_{2}}{2^{2}}+\ldots \dfrac{d_{m}}{2^{m}}\right) +z,
    \label{figure3}
\end{equation}
where $s$ is set to 0 and $z$ is the newly added zero point.
By freeing the 1-bit sign bit, which is ineffective in signed FP quantization, and using it as additional exponent / mantissa bit width, we fully utilize the representation capacity. As illustrated in Figure \ref{fig:unsign_mse}, unsigned FP quantization with a zero point significantly improves representation in over 95\% of AALs, compared to traditional signed FP quantization.

However, there are rare cases where performance slightly goes worse due to the diversity of anomalous distributions. Panel (c) of Figure \ref{fig:3_activations} shows that in such cases, the activation distribution may resemble a normal distribution, where signed FP quantization might perform better. 
Figure \ref{fig:unsign_mse} further indicates that introducing a zero point into signed FP quantization is unnecessary, offering minimal improvement in a few cases.

Given the strong performance of unsigned FP quantization with a zero point and the diversity of AALs, we propose a mixup-sign FP quantization framework. During the search-based initialization phase, we use signed FP quantization for NALs and introduce both unsigned FP quantization with a zero point and signed FP quantization for AALs. This approach addresses AAL challenges in low-bit quantization while minimizing computational overhead by adding the zero point only to unsigned FP quantization.

\subsection{Timestep-Aware Router for LoRA Allocation}
In Section~\ref{section3.2}, we observe that a single LoRA cannot fully capture all the information across timesteps due to the diverse generative characteristics at different timesteps. We also find that a reasonable allocation of different LoRAs for timesteps will be beneficial to fine-tuning. In this section, we introduce a timestep-aware LoRA allocation method that dynamically assigns optimal LoRA to each timestep, maximizing fine-tuning effectiveness.

Our method relies on a learnable router (illustrated in Figure~\ref{fig:pipeline}), a module shared across all timesteps. It takes the timestep as input and outputs selection probabilities for each LoRA across UNet layers. For each timestep, the LoRA with the highest probability corresponding to the router’s output is inserted into the quantized model for fine-tuning or inference. In a router network, the main components are a time embedding layer and an MLP layer. The time embedding layer, derived from a pre-trained diffusion model, converts the scalar into a $d$-dimensional embedding. The MLP layer then maps the embedding to a LoRA allocation distribution, where $y$ is the number of quantized layers and $h$ is the size of the LoRA Hub. Using an STE method~\cite{bengio2013estimating}, this distribution is converted into 0/1 probabilities to allocate suitable LoRAs to the diffusion model’s quantized layers.






\subsection{Denoising Factor Aligned Loss}

To address the mismatch between the actual performance gap and the loss used during quantization, we review the denoising principle depicted in Equation \ref{equation3} and pinpoint the cause of the mismatch: the time-step-dependent constraint on the predicted noise is not sufficiently accounted for during the denoising process. Therefore, we implement a modification to the loss function based on predicted noise:

\begin{equation}
    L^{t}=\gamma _{t}\cdot L_{\boldsymbol{\varepsilon} _{\theta }}^{t}.
\end{equation}

By introducing $\gamma_{t}$, which accurately reflects the utilization of the predicted noise at each time step, we achieve a preliminary alignment between the loss and the actual quantization error, as shown in Figure \ref{fig:3_loss}. This facilitates more accurate fine-tuning, leading to better performance recovery in the low-bit diffusion model.

%% file: sec/5_experiment.tex
\section{Experiment}
\subsection{Experimental Setup}

\textbf{Models and Metrics.} To verify the effectiveness of the proposed method, we evaluate it with two widely adopted diffusion paradigms: DDIM~\cite{song2020denoising} and LDM~\cite{rombach2022high}. For DDIM experiments, we evaluate on CIFAR-10 \cite{krizhevsky2009learning} and CelebA \cite{liu2015deep}. For LDM, we test unconditional generation on LSUN-Bedroom~\cite{yu2015lsun} and LSUN-Church~\cite{yu2015lsun} and conditional generation on ImageNet~\cite{deng2009imagenet}. The performance of the diffusion models is evaluated with Inception Score (IS) \cite{salimans2016improved}, Fréchet Inception Distance (FID) \cite{heusel2017gans} and Sliding FID (sFID)~\cite{salimans2016improved}. All metrics are evaluated based on 50k samples generated by the DDIM solver~\cite{song2020denoising}.

\textbf{Quantization Detail.} We employ standard layer-wise quantization for both weights and activations. Except for the input and output layers, which are typically set to 8-bit, all other convolution and linear layers are quantized to the target bit-width.
Furthermore, we generate 256 samples for the calibration set based on Q-Diffusion~\cite{li2023q} for bias initialization and use the method from~\cite{chen2024low} to obtain the optimal quantization parameters. 


\textbf{Baseline.} 
We compare two main types of quantization methods: PTQ methods (Q-Diffusion~\cite{ li2023q} and EDA-DM~\cite{ wang2023towards}) and fine-tuning methods (EfficientDM~\cite{he2023efficientdm} and QuEST~\cite{wang2024quest}). Since these fine-tuning methods involve special settings like non-full-layer quantization, we standardize the settings of EfficientDM in a consistent manner and procure other standardized results from DilateQuant~\cite{liu2024dilatequant}. Comparison with special settings is provided in Appendix.


\begin{table}[t]
\centering
\small
\renewcommand{\arraystretch}{1}
\setlength{\tabcolsep}{2.5mm}{
\begin{tabular}{@{}ccccc@{}}
\toprule
\midrule
\textbf{Task} & \textbf{Method}       & \makecell{\textbf{Prec.} \\ \textbf{(W/A)}} & \textbf{FID $\downarrow$} & \textbf{IS $\uparrow$} \\ \midrule\midrule
\multirow{11}{*}{\makecell{CIFAR-10 \\ 32x32 \\ \\ DDIM \\ steps = 100}}& FP                    & 32/32                & 4.26           & 9.03          \\ \cmidrule(lr){2-5}
& Q-Diffusion              & 6/6                  & 9.19          & 8.76 \\
&  EDA-DM       & 6/6                  & 26.68          & \textbf{9.35}          \\
& EfficientDM           & 6/6                  & 25.03          & 8.08          \\ 
& \cellcolor{green!20} Ours ($h$=2)  & \cellcolor{green!20} 6/6                  & \cellcolor{green!20} 4.26  & \cellcolor{green!20} 9.04          \\ 
& \cellcolor{blue!20} Ours ($h$=4)   & \cellcolor{blue!20}  6/6                  & \cellcolor{blue!20}  \textbf{4.23}             & \cellcolor{blue!20}  9.06           \\ \cmidrule(lr){2-5}
& Q-Diffusion               & 4/4                  & N/A         & N/A          \\
& EDA-DM          & 4/4                  & 120.24          & 4.42         \\
& EfficientDM           & 4/4                  & 38.40           & 7.32          \\ 
& \cellcolor{green!20} Ours ($h$=2)  & \cellcolor{green!20} 4/4                  & \cellcolor{green!20} \textbf{6.02}  & \cellcolor{green!20} 8.79          \\ 
& \cellcolor{blue!20} Ours ($h$=4)   & \cellcolor{blue!20}  4/4                  & \cellcolor{blue!20}  6.10           & \cellcolor{blue!20}  \textbf{8.90} \\ \midrule\midrule

\multirow{13}{*}{\makecell{LSUN \\ (Bedroom) \\ 256x256 \\ \\ LDM-4 \\ steps = 100 \\ eta = 1.0}}& FP                    & 32/32                & 3.02           & 2.29          \\ \cmidrule(lr){2-5}
& Q-Diffusion               & 6/6                  & 10.10          & 2.11          \\
& EDA-DM          & 6/6                  & 10.56           & 2.12          \\
& QuEST                & 6/6                  & 10.10           & 2.20          \\ 
& EfficientDM          & 6/6                  & 12.95  & \textbf{2.57}          \\ 
& \cellcolor{green!20}  Ours ($h$=2)  & \cellcolor{green!20}  6/6                  & \cellcolor{green!20}  8.42           & \cellcolor{green!20} 2.49\\ 
& \cellcolor{blue!20}   Ours ($h$=4) & \cellcolor{blue!20}   6/6                  & \cellcolor{blue!20}   \textbf{8.40}           & \cellcolor{blue!20}  2.49\\ \cmidrule(lr){2-5}
& Q-Diffusion               & 4/4                  & N/A            & N/A           \\
& EDA-DM          & 4/4                  & N/A          & N/A          \\
& QuEST                & 4/4                  & N/A            & N/A           \\ 
& EfficientDM           & 4/4                  &  36.36  & \textbf{2.69}          \\ 
& \cellcolor{green!20}  Ours ($h$=2)  & \cellcolor{green!20}  4/4                  & \cellcolor{green!20}  \textbf{12.21}          & \cellcolor{green!20}  2.47\\ 
& \cellcolor{blue!20}   Ours ($h$=4)  & \cellcolor{blue!20}   4/4                  & \cellcolor{blue!20}   12.34          & \cellcolor{blue!20}  2.48 \\ \midrule\midrule

\multirow{13}{*}{\makecell{LSUN \\ (Church) \\ 256x256 \\ \\ LDM-8 \\ steps = 100 \\ eta = 0.0}}& FP                    & 32/32                & 4.06           & 2.70          \\ \cmidrule(lr){2-5}
& Q-Diffusion               & 6/6                  & 10.90          & 2.47          \\
& EDA-DM          & 6/6                  & 10.76           & 2.43          \\
& QuEST                & 6/6                  & 6.83           & 2.65          \\ 
& EfficientDM           & 6/6                  & 7.45  & \textbf{2.80}          \\ 
& \cellcolor{green!20}  Ours ($h$=2)  & \cellcolor{green!20}  6/6                  & \cellcolor{green!20}  \textbf{6.24}           & \cellcolor{green!20}  2.73 \\ 
& \cellcolor{blue!20}  Ours ($h$=4)  & \cellcolor{blue!20}   6/6                  & \cellcolor{blue!20}   6.38           & \cellcolor{blue!20}   2.73 \\ \cmidrule(lr){2-5}
& Q-Diffusion               & 4/4                  & N/A            & N/A           \\
& EDA-DM         & 4/4                  & N/A           & N/A           \\
& QuEST                & 4/4                  & 13.03          & 2.63          \\ 
& EfficientDM           & 4/4                  & 18.40          &\textbf{2.97}           \\ 
& \cellcolor{green!20}  Ours ($h$=2)  & \cellcolor{green!20}  4/4                  & \cellcolor{green!20}  8.81           & \cellcolor{green!20} 2.70          \\ 
& \cellcolor{blue!20}   Ours ($h$=4)  & \cellcolor{blue!20}   4/4                  & \cellcolor{blue!20}   \textbf{8.77}  & \cellcolor{blue!20}   2.71  \\ \midrule\midrule
\end{tabular}
}
\caption{Quantization performance of unconditional generation. `Prec. (W/A)' denotes the quantization bit-width. `N/A' denotes failed image generation. $h$ denotes the size of LoRA Hub.}
\vspace{-10pt}
\label{table2}
\end{table}

\subsection{Quantization Performance}
\textbf{Unconditional Generation.}
Table~\ref{table2} presents the results of unconditional image generation across multiple datasets. With 6-bit quantization, our method achieves nearly identical performance to full precision. Our 4-bit quantized models achieve SOTA results in all tasks, significantly outperforming previous baseline methods. Notably, on CIFAR-10, our 4-bit quantized model results in an FID score that is only 1.84 worse than full precision, an almost negligible degradation, while previous methods struggle with 4-bit quantization.
Compared to the fine-tuning method \cite{he2023efficientdm}, our 4-bit quantized models improve FID by 32.38, 24.15, and 9.63 on three datasets, respectively, and still maintain remarkable IS. Additionally, we provide the performance of 4-bit and 6-bit quantized models on CelebA in Appendix.

\textbf{Conditional Generation.}
Table~\ref{table3} presents the results of our conditional generation experiments on ImageNet. We observe that FID is not always a reliable metric in this context, as an unexpected trend emerges: as the bit-width of the quantized model decreases, the FID score improves, which contradicts the expected trend. Therefore, our discussion of the ImageNet results focuses on other evaluation metrics. As shown by the IS and sFID, our method achieves performance comparable to that of the full-precision model with 6-bit quantization. Even with 4-bit quantization, we achieve SOTA results in terms of sFID, improving by 7.00 over the previous best method, EfficientDM~\cite{he2023efficientdm}. Furthermore, our method significantly outperforms two other fine-tuning methods in the IS metric. Visual evaluations further confirm that the generated images maintain high quality, exhibiting clear and coherent content, as shown in Figure~\ref{fig:visualization}. More visualization results will be presented in Appendix.

\begin{figure}[t]
    \centering
    \includegraphics[width=1\linewidth]{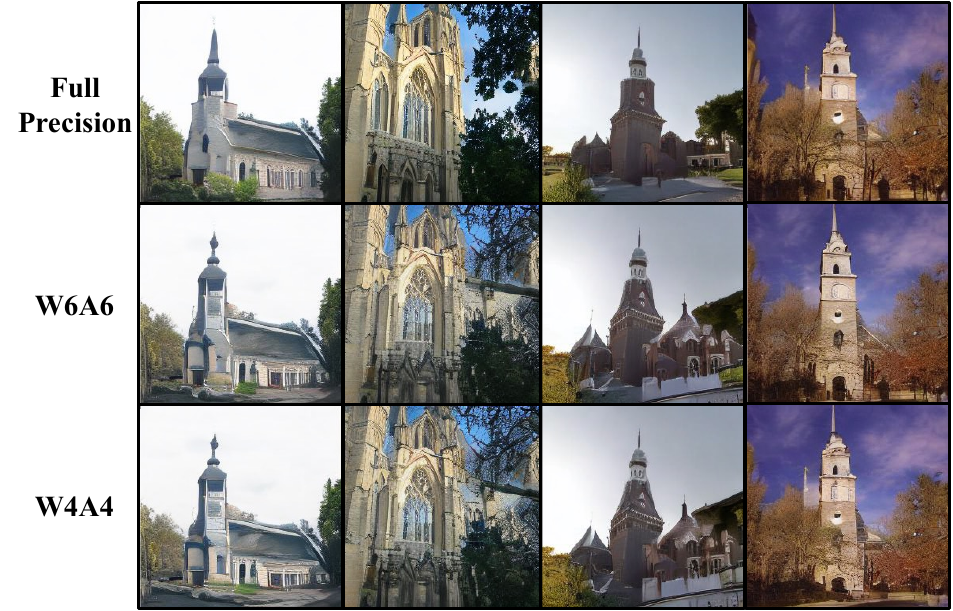}
    \caption{A visual comparison of generation results using our method across different quantization bit-widths, with the LSUN-Church dataset as an example.}
    \label{fig:visualization}
    \vspace{-10pt}
\end{figure}

\begin{table}[h]
\centering
\small
\begin{tabular}{@{}ccccc@{}}
\toprule
\textbf{Method}       & \textbf{Prec. (W/A)} & \textbf{sFID ↓} & \textbf{FID ↓} & \textbf{IS ↑} \\ \midrule
FP                    & 32/32                & 7.67            & 11.69          & 364.72        \\ \cmidrule(lr){1-5}
EDA-DM& 6/6                  & 8.02            & 11.52          & \textbf{360.77}        \\
QuEST& 6/6                  & 9.36            & \textbf{8.45}           & 310.12        \\ 
EfficientDM& 6/6                  & 6.88   & 9.54  & 351.79        \\ 
\cellcolor{green!20}  Ours ($h$=2)  & \cellcolor{green!20}6/6                  & \cellcolor{green!20}7.43            & \cellcolor{green!20}10.10          & \cellcolor{green!20}349.91        \\ 
\cellcolor{blue!20}   Ours ($h$=4)  & \cellcolor{blue!20}6/6                  & \cellcolor{blue!20}\textbf{6.65}            & \cellcolor{blue!20}10.10          & \cellcolor{blue!20}351.79 \\ \cmidrule(lr){1-5}

EDA-DM& 4/4                  & 36.66           & 20.02          & \textbf{204.93}        \\
QuEST& 4/4                  & 29.27           & 38.43          & 69.58         \\ 
EfficientDM& 4/4                  & 14.42           & 12.73           & 139.45 \\ 
\cellcolor{green!20}  Ours ($h$=2)  & \cellcolor{green!20}4/4                  & \cellcolor{green!20}\textbf{7.42}& \cellcolor{green!20}\textbf{6.50}  & \cellcolor{green!20}190.74        \\ 
\cellcolor{blue!20}   Ours ($h$=4)  & \cellcolor{blue!20}4/4                  & \cellcolor{blue!20}8.23   &\cellcolor{blue!20} 7.43           & \cellcolor{blue!20}177.40  \\ \midrule
\end{tabular}
\vspace{-5pt}
\caption{Quantization performance of conditional generation for fully-quantized LDM-4 models on ImageNet 256×256 with 20 steps. `Prec. (W/A)' denotes the quantization bit-width. $h$ denotes the size of LoRA Hub. }
\label{table3}
\vspace{-5pt}
\end{table}

\subsection{Ablation Study}
\begin{figure}[h]
    \centering
    \includegraphics[width=0.9\linewidth]{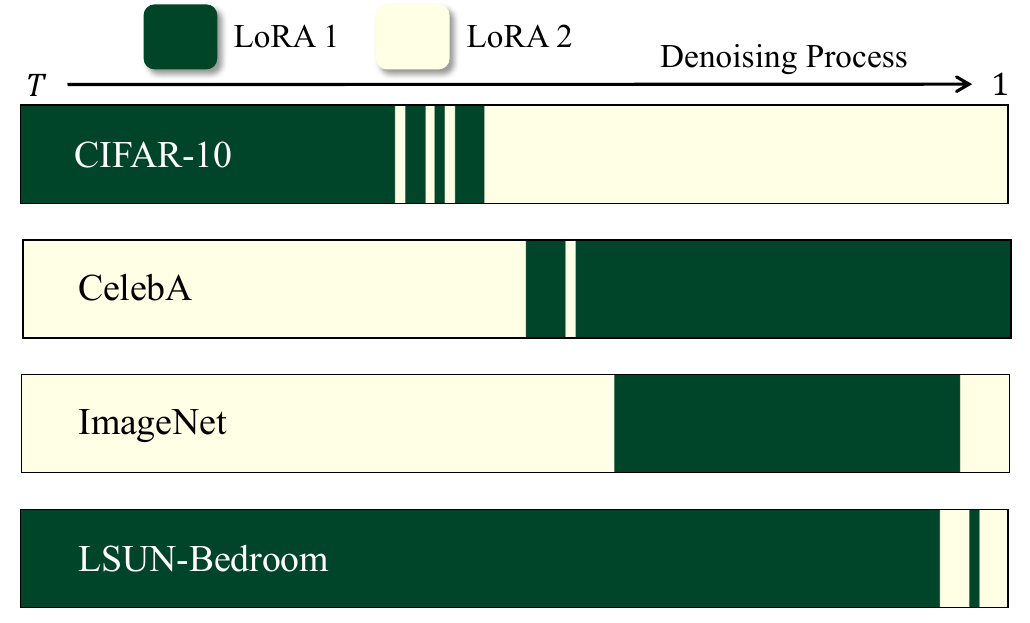}
    \caption{Distribution of LoRA allocations over timesteps obtained after router training on different datasets, when $h=2$.}
    \label{fig:router}
    \vspace{-15pt}
\end{figure}

The ablation experiments are conducted on the 4-bit quantization using the CelebA dataset, which is challenging for low-bit quantization, further demonstrating the effectiveness of our approach. 
The baseline uses signed FP quantization combined with single LoRA fine-tuning. As shown in Table \ref{table4}, all three proposed modules lead to significant performance improvements, with their combination yielding a synergistic effect. The baseline FID score is 9.53 higher than that of the full-precision model (6.49). By applying our technique, we reduce the FID by 8.18 compared to the baseline.
More interesting is that we visualize the LoRA allocation distribution learned by the router as shown in Figure~\ref{fig:router}. 
We find that the distribution of the router-learned allocation over timesteps is consistent with the finding that the diffusion model focuses on contour generation early and on detail generation later~\cite{wang2023diffusion}.

\begin{table}[t]
\centering
\small
\setlength{\tabcolsep}{4.mm}{
\begin{tabular}{@{}ccc|cc@{}}
\toprule
 \multicolumn{3}{c|}{\textbf{Method}} & \multirow{2}{*}{\makecell{\textbf{Prec.} \\ \textbf{(W/A)}}}& \multirow{2}{*}{\makecell{\textbf{FID $\downarrow$} }}\\
MSFP               & TALoRA    & DFA    &  &        \\ 
\midrule

\XSolidBrush                & \XSolidBrush          & \XSolidBrush    &  4/4&16.02                \\
\Checkmark                  & \XSolidBrush          & \XSolidBrush    &  4/4&9.60                 \\
\XSolidBrush                   & \Checkmark          & \XSolidBrush    &  4/4
&10.66                 \\ 
\Checkmark                  & \XSolidBrush          & \Checkmark      &  4/4&8.39                 \\ 
\Checkmark                  & \Checkmark            & \XSolidBrush    &  4/4
&8.79                 \\ 
\Checkmark                  & \Checkmark            & \Checkmark      &  4/4&7.69                 \\ 
\bottomrule

\end{tabular}}
\caption{Ablation study on different modules we proposed. Testing on CelebA dataset with $h=2$ LoRA Hub size.}
\vspace{-15pt}
\label{table4}

\end{table}

%% file: sec/6_conclusion.tex
\section{Conclusion}
In this paper, we focus on exploring low-bit FP quantization for diffusion models. For model initialization, we innovatively introduce unsigned FP quantization with zero point to address AALs. For the fine-tuning based on the denoising process, we formulate it as a multi-task procedure. We introduce multiple LoRAs along with a router for their allocation at different timesteps, and further align the loss function, originally based on estimated noise, with the actual quantization error. We introduce unsigned FP quantization and achieve 4-bit FP quantized diffusion models. Our FP PTQ-based fine-tuning method sets a new precedent for 4-bit diffusion models, offering insight into the deployment of low-bit diffusion models in the future.

%% file: sec/7_acknowledge.tex
\section{Acknowledgments}
This work is supported by Shanghai Science and Technology Commission Explorer Program Project (24TS1401300), National Key Research and Development Program of China (No.2022ZD0160101).
The computations in this research were performed using the CFFF platform of Fudan University.

%% file: sec/X_suppl.tex
\clearpage
\appendix
\setcounter{page}{1}
\maketitlesupplementary

\section{Supplementary Material Overview}
\label{sec:overview}
In this supplementary material, we provide additional explanations and experimental results referenced in the main paper. The content is organized as follows:
\begin{itemize}
    \item Methodology of Mixup-Sign Quantization in Appendix \ref{Methodology of Mixup-sign Quantization}.
    \item More Implementation Details in Appendix
    \ref{More Implementation Details}.
    \item FP vs. INT in Post-Training Quantization  in Appendix
    \ref{FP vs. INT in Post-Training Quantization}.
    \item Comprehensive Analysis of TALoRA Performance in Appendix
    \ref{Comprehensive Analysis of TALoRA Performance}.
    \item Supplementary Performance Evaluation in Appendix
    \ref{Supplementary Performance Evaluation}.
    \item Extensive Comparison with EfficientDM and QUEST in Appendix
    \ref{Extensive Comparison with EfficientDM and QUEST}.
    \item Additional Visualization Results in Appendix
    \ref{Additional Visualization Results}.
\end{itemize}

\section{Methodology of Mixup-Sign Quantization}
\label{Methodology of Mixup-sign Quantization}
We implement the proposed MSFP strategy using a search-based method~\cite{liu2023llm,chen2024low}, wherein the quantization parameters are determined by minimizing the MSE between the distributions before and after quantization. 

To clarify, the quantization parameters for the signed FP quantization include the $format$, bias $b$, and sign bit $s$ set to 1, whereas the quantization parameters for the unsigned FP quantization include the $format$, bias $b$, sign bit $s$ set to 0, and zero point $zp$. All quantization parameters are assigned a search space during initialization.

As mentioned in the main text, the bias $b$ serves as a threshold in FP quantization:

\begin{equation}
    maxval = 2^{2^{x}-1-b}\cdot \left( 1-\dfrac{1}{2^{y}}\right) 
\end{equation}

The maximum value, denoted as $maxval$, is determined by the $format$ (e.g.,ExMy) abd the bias $b$, and represents the maximum discrete value achievable in FP quantization. Notably, $maxval$ and $b$ are directly correlated, and for convenience, we will refer to $maxval$ in subsequent discussions as the equivalent to the bias $b$.

\begin{algorithm}[t]
    \caption{Initialization of Quantization Parameters} \label{MSFP}
    \small
    \begin{algorithmic}[1]
     \STATE \textbf{Input:} $format\_options$, $maxval\_options$, $(zp\_options)$, $(unsigned\_format\_options)$
     \STATE \textbf{Output:} $format$, $maxval$, $(zp)$
     \STATE
     \STATE \textit{\#10000 is huge enough}
     \STATE $min\_mse = 10000$ 
     \STATE $s=1$
     \FOR{$f$ in $format\_options$}
        \FOR{$prev\_m$ in $maxval\_options$}
            \STATE $prev\_mse = calculate\_mse(f,prev\_m,s)$
            \IF{$prev\_mse<min\_mse$}
                \STATE $min\_mse = prev\_mse$
                \STATE $format = f$
                \STATE $maxval = prev\_m$
            \ENDIF
        \ENDFOR
     \ENDFOR
     \STATE
     \STATE \textit{\#only for unsigned FP quantization}
     \STATE $s=0$
     \FOR{$f$ in $unsigned\_format\_options$}
        \FOR{$prev\_m$ in $maxval\_options$}
            \FOR{$prev\_zp$ in $zp\_options$}
                \STATE $prev\_mse =$ \\
                $calculate\_mse(f,prev\_m,prev\_zp,s)$
                \IF{$prev\_mse<min\_mse$}
                    \STATE $min\_mse = prev\_mse$
                    \STATE $format = f$
                    \STATE $maxval = prev\_m$
                    \STATE $zp = prev\_zp$
                \ENDIF
            
            \ENDFOR
        \ENDFOR
    \ENDFOR
        
    \end{algorithmic}
\end{algorithm}

In the MSFP strategy, initialization is divided into two parts: weight initialization and activation initialization. During initialization, we determine the optimal quantization parameter settings, and the process is outlined in Algorithm \ref{MSFP}: In the first stage, the search for signed FP quantization parameters is applicable to all cases. In the second stage, the search for unsigned FP quantization parameters is specifically applied to the activation initialization of the Anomalous-Activation-Distribution Layers (AALs) mentioned in the main text.

Additionally, due to the significant variability in the search space for $maxval$, which depends on the differing distributions of the data, therefore, prior to initiating the search for quantizer parameters, the first step involves performing several random forward passes to capture the maximum value observed for each quantizer. This value is then used as the initial $maxval\_0$.

\begin{figure*}[htbp]
    \centering
    \includegraphics[width=1\linewidth]{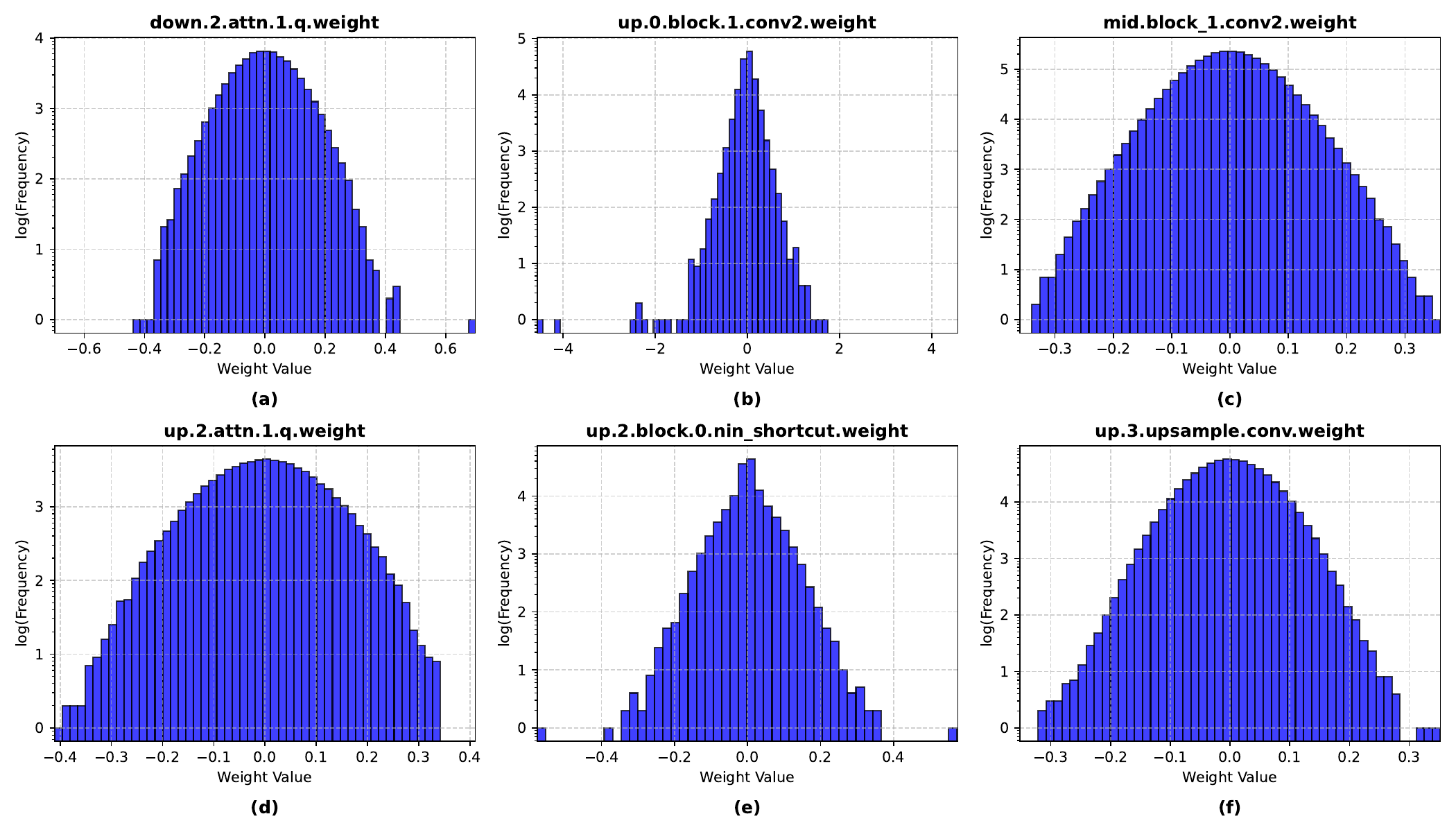}
    \caption{The weight distribution of certain layers in the DDIM model on CelebA dataset.}
    \label{fig:6W}
\end{figure*}

\textbf{Weight Initialization.} For weight initialization, since the distribution of weights typically approximates a normal distribution (as shown in Figure \ref{fig:6W}), we deploy signed FP quantization. In the search for the $format$ of signed FP quantization, we define a search space of size 4 for 4-bit, 6-bit, and 8-bit representations, encompassing the most expressive data formats for each bit-width while striking a balance between computational overhead and performance~\cite{kuzmin2022fp8,micikevicius2022fp8,van2023fp8}.

\begin{table}
\centering
\begin{tabular}{ccc}
\toprule
\textbf{Search Space}         & \textbf{\begin{tabular}[c]{@{}c@{}}Bits (W/A)\end{tabular}} & \textbf{FID $\downarrow$} \\
\midrule
{[}0,maxval\_0{]}             & 6/32                                                           & 10.14        \\
{[}0,2maxval\_0{]}            & 6/32                                                           &   10.26           \\
{[}0.6maxval\_0,2maxval\_0{]} & 6/32                                                           & 9.36         \\
{[}0.7maxval\_0,2maxval\_0{]} & 6/32                                                           & 6.46         \\
{[}0.8maxval\_0,2maxval\_0{]} & 6/32                                                           & 5.58         \\
{[}0.9maxval\_0,2maxval\_0{]} & 6/32                                                           & \textbf{5.13}         \\
{[}maxval\_0,2maxval\_0{]}    & 6/32                                                           & 5.83        \\
\bottomrule   
\end{tabular}
\caption{The impact of different $maxval$ search spaces in weight initialization on the DDIM model performance on CelebA dataset.}
\label{Weight_maxval}
\vspace{-10pt}
\end{table}

For the search of $maxval$ in weights, we extend the previous search range of $range(0,maxval\_0,0.001)$ to explore a more refined and reasonable search space. On the one hand, considering that large-value weights are relatively few but have a significant impact, we set the lower bound of the search to a value slightly smaller than $maxval\_0$ to avoid excessive loss of essential large-value weights. On the other hand, setting the upper bound to $maxval\_0$ may not guarantee the minimization of MSE. As inferred from the representation of FP quantization, any quantizer with its $maxval$ larger than $2\times maxval\_0$ cannot result in a smaller MSE, so we set the upper bound of the search to $2\times maxval\_0$. As shown in Table \ref{Weight_maxval}, our exploration across different search spaces demonstrates the effectiveness of the redefined search space of $maxval$. 

\textbf{Activation Initialization.} 
For activation initialization, based on the analysis in the main text, we employ signed FP quantization for NALs with distribution approximately following a normal distribution, and adopt a mixup-sign FP quantization strategy for AALs with asymmetric distributions. Unlike weight initialization, where weights remain static, activation initialization needs to account for potential activation distributions. To ensure that the activations used for initialization are representative, we introduce a calibration dataset~\cite{shang2023post,li2023q}, as is common in INT quantization.

Given the increased complexity and randomness of activation distributions, we include all possible formats for different bit-widths within the search space for $format$. Notably, for $n$-bit unsigned FP quantization with the ExMy $format$, the condition $x + y+s=n$ applies, where $s=0$, distinguishing its format from that of signed FP quantization, which includes $s$ set to 1 under the same bit-width. Accordingly, the search range for $maxval$ is adjusted to $linspace(0,maxval\_0,100)$, preventing excessive computational overhead. Lastly, for the zero point $zp$ introduced in unsigned FP quantization, since the minimum value of the distribution is constrained by $SiLU$ to approximately -0.278, assigning $zp$ a search space of $linsapce(-0.3,0,6)$ is sufficient.

\section{More Implementation Details}
\label{More Implementation Details}
\textbf{FP PTQ Configuration.}
Following the procedure outlined in Appendix \ref{Methodology of Mixup-sign Quantization}, we deploy our MSFP strategy for both weights and activations. The initialization of $maxval\_0$ is achieved by generating 2000 images through random forward passes. Subsequently, a calibration dataset is constructed based on the output of the full-precision model, following the approach of Q-Diffusion~\cite{li2023q}. Specifically, 256 samples are used for the DDIM model, while 128 samples are used for the LDM model.

For weight initialization, the search spaces for $maxval$ and $format$ are presented in Table \ref{Weight_parameters}. For activation initialization, the search spaces for $maxval$, $fomrat$ and $zp$ are thoroughly discussed and provided in Appendix \ref{Methodology of Mixup-sign Quantization}.


\begin{table}[h!]
\centering
\small
\begin{tabular}{ccc}
\toprule
\textbf{Bit} & \textbf{\begin{tabular}[c]{@{}c@{}}Search Space\\ ($maxval$)\end{tabular}} & \textbf{\begin{tabular}[c]{@{}c@{}}Search Space\\ ($format$)\end{tabular}} \\
\midrule
4            & {[}0.8maxval\_0,2maxval\_0{]}                                            & {[}E3M0,E2M1,E1M2,E0M3{]}                                                \\
6            & {[}0.9maxval\_0,2maxval\_0{]}                                            & {[}E4M1,E3M2,E2M3,E1M4{]}                                                \\
8            & {[}0.9maxval\_0,2maxval\_0{]}                                            & {[}E5M2,E4M3,E3M4,E2M5{]}     \\                              
\bottomrule
\end{tabular}
\caption{Search spaces for different quantization parameters under different bit-widths in weight initialization.}
\label{Weight_parameters}
\end{table}

\textbf{Fine-tuning Configuration.} For the noise estimation U-Net, all quantized layers, except for the input and output layers, are quantized and equipped with QLoRA-based TALoRAs~\cite{dettmers2023qlora}. Each TALoRA is initialized with a rank of 32. The selection of different TALoRAs at each timestep is managed by a router, which is implemented as a linear layer. The input channels of the router match the channel count of the timestep embedding in the diffusion model. 

Adam optimizers are assigned to both the TALoRAs and the router, with a learning rate of 1e-4 for both components. Fine-tuning is performed for 160 epochs with a batch size of 16 on DDIM models and 320 epochs with a batch size of 8 on LDM models. Notably, the batch size for the ImageNet dataset is reduced to 4.


\section{FP vs. INT in Post-Training Quantization}
\label{FP vs. INT in Post-Training Quantization}
Table \ref{MSFP-ONLY} presents a performance comparison between the 6-bit model initialized with MSFP and several 6-bit models based on traditional INT quantization~\cite{shang2023post,li2023q,wang2023towards,esserlearned}. As shown, even without fine-tuning, our approach significantly outperforms existing SOTA methods in handling 6-bit quantization for diffusion models. This highlights that FP quantization is a more effective choice for handling low-bit activation quantization in diffusion models, a task that is both challenging and crucial, compared to INT-based methods.

\begin{table}[t]
\centering
\small
\renewcommand{\arraystretch}{1}
\begin{tabular}{@{}ccccc@{}}
\toprule
\midrule
\textbf{Task} & \textbf{Method}       & \makecell{\textbf{Prec.} \\ \textbf{(W/A)}} & \textbf{FID $\downarrow$} & \textbf{IS $\uparrow$} \\ \midrule\midrule
\multirow{6}{*}{\makecell{CelebA \\ 64x64 \\ \\ DDIM \\ steps = 100}}& FP                    & 32/32                & 6.49           & 2.61          \\ \cmidrule(lr){2-5}
& LSQ              & 6/6                  & 78.37          & 1.94 \\
&  PTQ4DM      & 6/6                  & 24.96          & 2.13          \\
& Q-Diffusion           & 6/6                  & 23.37          & 2.16          \\ 
& ADP-DM          & 6/6                  & 16.86          & 2.30
\\
& Ours(MSFP)           & 6/6                  & \textbf{9.51}         & \textbf{2.78} 
 \\ \midrule\midrule

\end{tabular}
\caption{Quantization performance of unconditional generation. In this case, 'Ours' refers to the method that deploys only the MSFP strategy without any fine-tuning. `Prec. (W/A)' denotes the quantization bit-width.}
\vspace{-10pt}
\label{MSFP-ONLY}
\end{table}

\section{Comprehensive Analysis of TALoRA Performance}
\label{Comprehensive Analysis of TALoRA Performance}
\subsection{TALoRA Outperforms Rank-Scaled LoRA}
In our approach, we introduce multiple TALoRAs for the majority of quantized layers, which leads to an increase in the model size. Some may question whether the observed performance improvement is simply due to the larger memory footprint of the LoRAs. However, in practice, only one TALoRA is active at each timestep, which differs fundamentally from using a larger-rank LoRA, as the latter would result in higher training and inference costs. Furthermore, Table \ref{RANK_QLoRA} presents the results of fine-tuning with two TALoRAs (rank=32) and a single QLoRA (rank=64). Our method achieves even better performance, demonstrating that our timestep-aware fine-tuning strategy effectively recovers the performance lost during quantization in diffusion models, with lower overhead and enhanced performance.

\begin{table}[h!]
\centering
\begin{tabular}{cccc}
\toprule
\textbf{Method} & \textbf{Rank} & \textbf{Bits(W/A)} & \textbf{FID$\downarrow$} \\
\midrule
FP              & /             & 32/32              & 6.49         \\
single-LoRA    & 64            & 4/4                & 7.75         \\
TALoRA($h$=2)     & 32            & 4/4                & \textbf{7.69}        \\
\bottomrule
\end{tabular}
\caption{Comparison between TALoRA and rank-scaled LoRA in fine-tuning 4-bit DDIM models on CelebA dataset. 'Rank' refers to the LoRA rank.}
\label{RANK_QLoRA}
\end{table}

\subsection{Impact of TALoRA Quantity}
\begin{figure}
    \centering
    \includegraphics[width=1\linewidth]{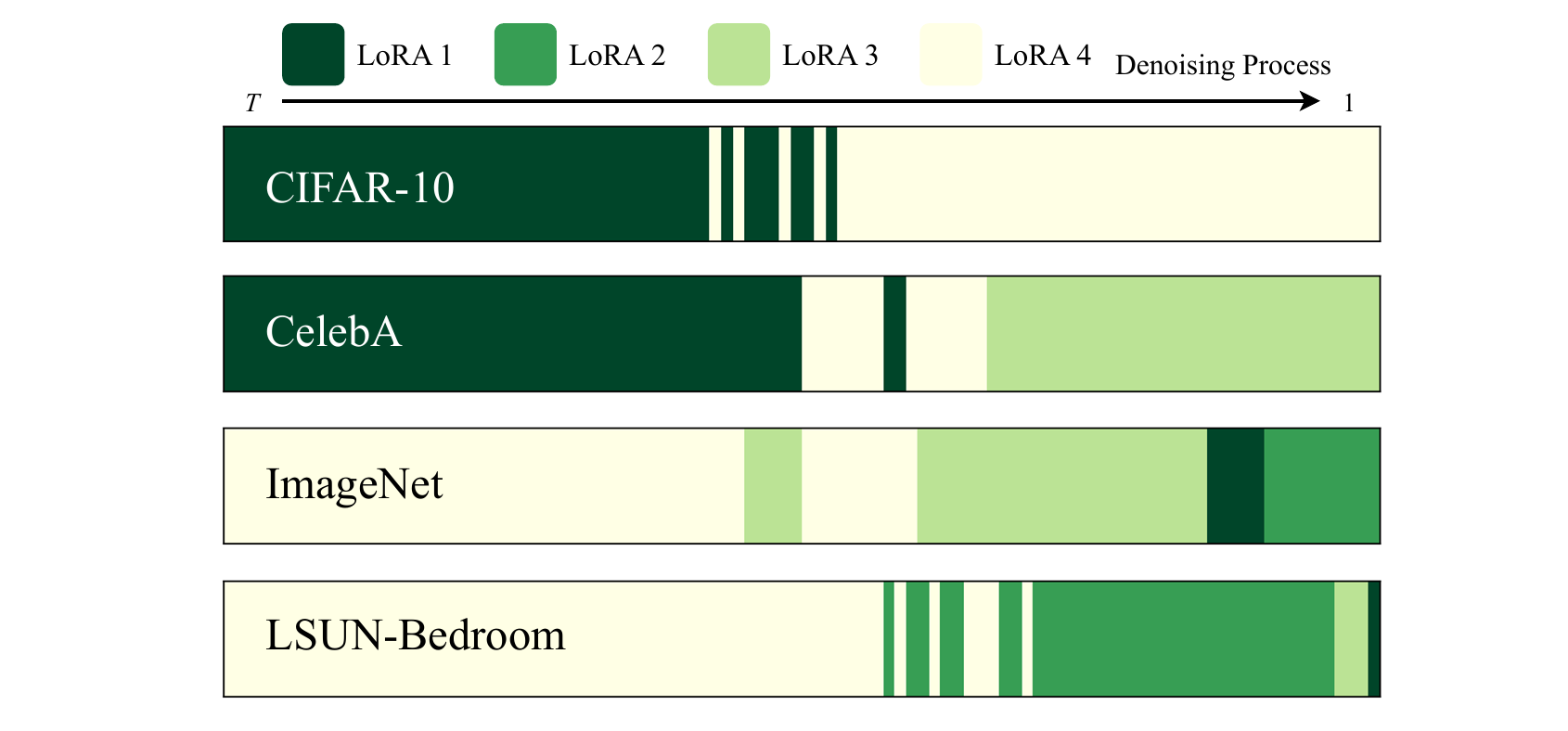}
    \caption{Distribution of LoRA allocations over timesteps obtained after router training on different datasets, when $h$ = 4.}
    \label{fig:4LoRA}
    \vspace{-10pt}
\end{figure}

As illustrated in Figure \ref{fig:4LoRA}, when deploying four TALoRAs, the distributions of LoRA allocation across different timesteps exhibits a strong regularity: in most cases, regardless of the dataset, the majority of timesteps utilize only two LoRAs. This suggests that fine-tuning low-bit diffusion models predominantly follows a two-stage task pattern, which aligns with the motivation behind introducing TALoRAs—viewing the denoising process as a progression from restoring coarse structures to refining intricate details~\cite{wang2023diffusion}. 

Experimental results in the main text further demonstrates that deploying four TALoRAs does not yield better results compared to deploying two TALoRAs. In fact, in most cases, the latter achieves superior results on 4-bit diffusion models. This aligns with our earlier analysis: two TALoRAs are sufficient to handle the fine-tuning task effectively, while the introduction of additional TALoRAs could reduce the training opportunities for the most impactful LoRAs, ultimately compromising fine-tuning performance.

\section{Supplementary Performance Evaluation}
\label{Supplementary Performance Evaluation}
To further validate the effectiveness of our approach, we conduct supplementary experiments. For the DDIM model, where prior methods have struggled, our approach is evaluated on the CelebA dataset~\cite{liu2015deep}—a more complex dataset with higher image resolutions corresponding to a more intricate DDIM model. As shown in Table \ref{CELEBA}, our method achieves cutting-edge performance under both 4-bit and 6-bit settings. Notably, our 4-bit diffusion model exhibits performance on FID and IS metrics comparable to full precision, and our method even outperforms the full-precision model under the 6-bit setting.

For the LDM model, we further evaluate it on the ImageNet dataset~\cite{deng2009imagenet} using two advanced sampling methods, PLMS~\cite{liu2022pseudo} and DPM-Solver~\cite{lu2022dpm}, which are more sophisticated and computationally demanding during fine-tuning. Table \ref{ImageNet} demonstrates that our method maintains robust performance under both 4-bit and 6-bit quantization settings, achieving SOTA results on the more reliable sFID and IS metrics in ImageNet.

Furthermore, we apply our method to the task of quantizing text-to-image diffusion models, specifically deploying it on Stable Diffusion with the MS-COCO dataset~\cite{lin2014microsoft}. Our approach also delivers highly satisfactory results, with detailed visualizations provided in Appendix \ref{Additional Visualization Results}.

\begin{table}[t]
\centering
\small
\renewcommand{\arraystretch}{1}
\begin{tabular}{@{}ccccc@{}}
\toprule
\midrule
\textbf{Task} & \textbf{Method}       & \makecell{\textbf{Prec.} \\ \textbf{(W/A)}} & \textbf{FID $\downarrow$} & \textbf{IS $\uparrow$} \\ \midrule\midrule
\multirow{9}{*}{\makecell{CelebA \\ 64x64 \\ \\ DDIM \\ steps = 100}}& FP                    & 32/32                & 6.49           & 2.61          \\ \cmidrule(lr){2-5}
& Q-Diffusion           & 6/6                  & 23.37          & 2.16          \\ 
& ADP-DM          & 6/6                  & 16.86          & 2.30
\\
& \cellcolor{green!20}Ours($h$=2)           & \cellcolor{green!20}6/6                  &\cellcolor{green!20} 5.38         & \cellcolor{green!20}\textbf{2.67} \\
&\cellcolor{blue!20} Ours($h$=4)           & \cellcolor{blue!20}6/6                  & \cellcolor{blue!20}\textbf{5.36}         & \cellcolor{blue!20}2.66
\\ \cmidrule(lr){2-5}
& Q-Diffusion           & 4/4                  & N/A          & N/A          \\ 
& ADP-DM          & 4/4                  & N/A          & N/A
\\
& \cellcolor{green!20}Ours($h$=2)           & \cellcolor{green!20}4/4                  & \cellcolor{green!20}\textbf{7.69}         & \cellcolor{green!20}2.59 \\
& \cellcolor{blue!20}Ours($h$=4)           &\cellcolor{blue!20} 4/4                  &\cellcolor{blue!20} 7.84         & \cellcolor{blue!20}\textbf{2.60}
 \\ \midrule\midrule

\end{tabular}
\caption{Quantization performance of unconditional generation. `Prec. (W/A)' denotes the quantization bit-width. ‘N/A’ denotes failed image generation. $h$ denotes the size of LoRA Hub.}
\vspace{0pt}
\label{CELEBA}
\end{table}

\begin{table}[t]
\centering
\small
\renewcommand{\arraystretch}{1}
\begin{tabular}{@{}cccccc@{}}
\toprule
\midrule
\textbf{Task} & \textbf{Method}       & \makecell{\textbf{Prec.} \\ \textbf{(W/A)}} & \textbf{sFID $\downarrow$} & \textbf{FID $\downarrow$} & \textbf{IS $\uparrow$}   \\ \midrule\midrule
\multirow{10}{*}{\makecell{LDM-4 \\ \\PLMS \\steps = 20}}& FP                    & 32/32         &   7.08    &  11.71          & 379.19          \\ \cmidrule(lr){2-6}
& EDA-DM           & 6/6      &    6.59        & 11.27          & 363.00          \\ 
& EfficientDM          & 6/6         &  9.36   & \textbf{9.85}    & 325.13
\\
&\cellcolor{green!20} Ours($h$=2)           & \cellcolor{green!20}6/6        &  \cellcolor{green!20}  5.63      &\cellcolor{green!20} 10.35         &\cellcolor{green!20}363.79 \\
&\cellcolor{blue!20} Ours($h$=4)           &\cellcolor{blue!20} 6/6         &    \cellcolor{blue!20} \textbf{5.33}    &\cellcolor{blue!20}10.25        & \cellcolor{blue!20}\textbf{364.27}
\\ \cmidrule(lr){2-6}
& EDA-DM           & 4/4        &  32.63    & 17.56   & \textbf{203.15}         \\ 
& EfficientDM          & 4/4           &    9.89   & 14.78      & 103.34
\\
& \cellcolor{green!20}Ours($h$=2)           &\cellcolor{green!20} 4/4      &   \cellcolor{green!20} \textbf{7.39}  & \cellcolor{green!20}\textbf{7.27}      & \cellcolor{green!20}196.32 \\
& \cellcolor{blue!20}Ours($h$=4)           & \cellcolor{blue!20}4/4       &     \cellcolor{blue!20} 7.83     & \cellcolor{blue!20}7.83      & \cellcolor{blue!20}193.11
 \\ \midrule\midrule
\multirow{10}{*}{\makecell{LDM-4 \\ \\DPM-\\Solver \\steps = 20}}& FP                    & 32/32         &   6.85    &  11.44          & 373.12          \\ \cmidrule(lr){2-6}
& EDA-DM           & 6/6      &    7.95        & 11.14          & 357.16          \\ 
& EfficientDM          & 6/6         &  9.30   &\textbf{8.54}   & 336.11
\\
& \cellcolor{green!20}Ours($h$=2)           &\cellcolor{green!20} 6/6        &  \cellcolor{green!20}  \textbf{6.86}      &\cellcolor{green!20} 9.61         &\cellcolor{green!20}363.71 \\
& \cellcolor{blue!20}Ours($h$=4)           & \cellcolor{blue!20}6/6         &   \cellcolor{blue!20} 6.88    &\cellcolor{blue!20}9.59        & \cellcolor{blue!20}\textbf{364.30}
\\ \cmidrule(lr){2-6}
& EDA-DM           & 4/4        &  39.40    & 30.86   & 138.01         \\ 
& EfficientDM          & 4/4           &  13.82   & 14.36      & 109.52
\\
&\cellcolor{green!20} Ours($h$=2)           & \cellcolor{green!20}4/4      & \cellcolor{green!20} \textbf{12.61} & \cellcolor{green!20}\textbf{8.46}     & \cellcolor{green!20}\textbf{257.33}\\
& \cellcolor{blue!20}Ours($h$=4)           & \cellcolor{blue!20}4/4       &  \cellcolor{blue!20}14.56     &\cellcolor{blue!20} 9.64      & \cellcolor{blue!20}238.07
 \\ \midrule\midrule

\end{tabular}
\caption{Quantization performance of conditional generation for
fully-quantized LDM-4 models on ImageNet 256×256 with 20
steps, using PLMS and DPM-Solver as sampling methods. ‘Prec. (W/A)’ denotes the quantization bit-width. $h$ denotes the size of LoRA Hub.}
\vspace{-10pt}
\label{ImageNet}
\end{table}

\section{Extensive Comparison with EfficientDM and QUEST}
\label{Extensive Comparison with EfficientDM and QUEST}
As mentioned in the main text, prior fine-tuning-based methods, such as EfficientDM~\cite{he2023efficientdm} and Quest~\cite{wang2024quest}, adopt specialized experimental setups. EfficientDM retains all $skip\_connection$ layers and the $op$ layers within $Upsample$ blocks in full precision. These layers constitute a significant portion of the model, and their quantization significantly affects performance. Therefore, in our comparative experiments, we apply standard quantization to these layers. In contrast, Quest adopts a different strategy by modifying the quantization granularity for activations. Specifically, in low-bit quantization, channel-wise quantization of weights is a common approach. However, Quest extends this to activations, introducing substantial computational overhead compared to the mainstream layer-wise quantization. To ensure a fair comparison, we employ conventional layer-wise quantization for activations.
\begin{figure*}[hb]
    \centering
    \vspace{0pt}
    \includegraphics[width=1\linewidth]{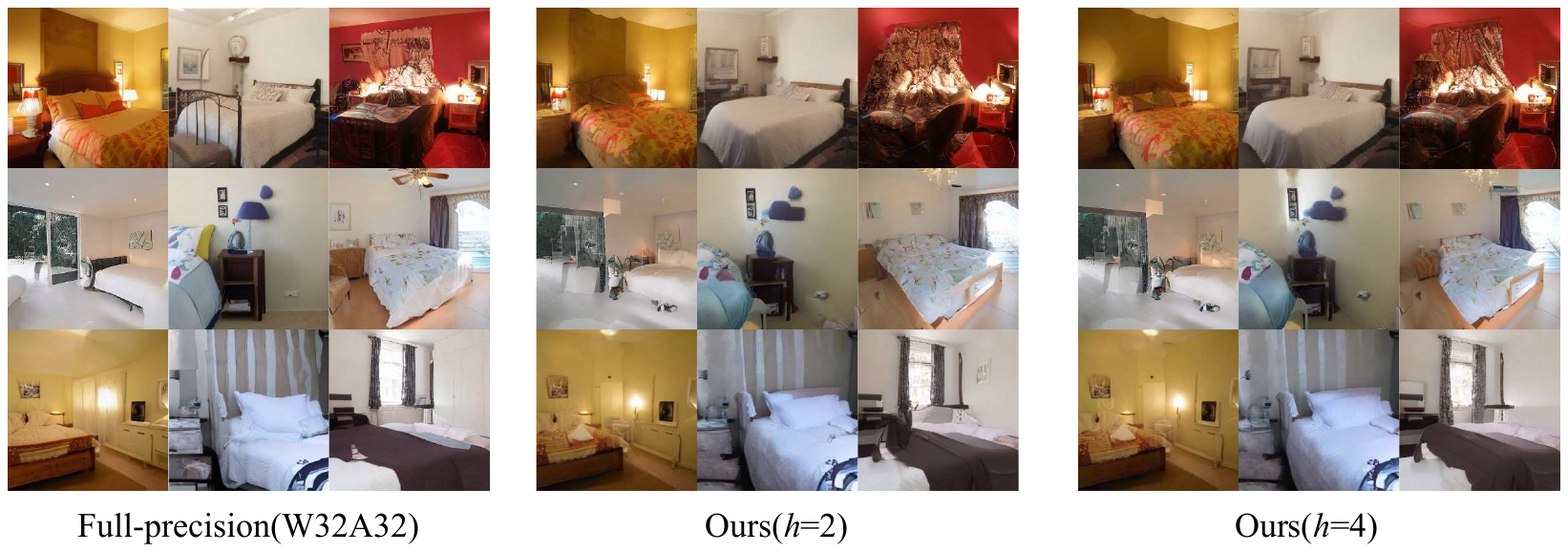}
    \vspace{-160pt} 
    \caption{Visualization of random samples from 4-bit LDM-4 on LSUN-Bedroom across different LoRA Hub sizes $h$.}
    \label{fig:Bedroom}
\end{figure*}

For a comprehensive evaluation, we align our method with the specific settings of EfficientDM and consider the implications of Quest's setup. As shown in Table \ref{Standard}, under EfficientDM's configuration, our 4-bit LDM model achieves significantly better results on the Church dataset, with an FID score that is 6.39 lower than EfficientDM’s. However, we choose not to replicate Quest's specific settings for two key reasons. First, despite using the more efficient layer-wise quantization for both weights and activations, our method already surpasses Quest’s performance. Specifically, under the 4-bit setting, our method achieves an FID of 8.81, compared to Quest’s 11.76, which relies on computationally expensive channel-wise quantization for both. Second, our approach relies on FP quantization, and incorporating channel-wise quantization necessitates search-based initialization for every channel, which is computationally infeasible.

\begin{table}[t]
    \centering
    \small
    \renewcommand{\arraystretch}{1}
    \begin{tabular}{@{}ccccc@{}}
    \toprule
    \midrule
    \textbf{Task} & \textbf{Settings}& \textbf{Method} & \makecell{\textbf{Prec.} \\ \textbf{(W/A)}} & \textbf{FID $\downarrow$} \\
    \midrule\midrule
    \multirow{10}{*}{\makecell{LSUN- \\Church \\ 256 $\times$ 256 \\ \\LDM-8 \\ steps = 100 \\ eta = 0.0}} 
    & - & FP & 32/32 & 4.06 \\ \cmidrule(lr){2-5}
    & \multirow{2}{*}{\makecell{\textbf{Partial} \\Quantization}} & EfficientDM & 4/4 & 13.68 \\
    & & Ours($h$=2) & 4/4 & \textbf{7.95} \\
    \cmidrule(lr){2-5}
    & \multirow{2}{*}{\makecell{\textbf{Full} \\ Quantization}} & EfficientDM & 4/4 &  18.40\\
    & & Ours($h$=2) & 4/4 & \textbf{8.81} \\
    \cmidrule(lr){2-5}
    & \multirow{2}{*}{\makecell{\textbf{Channel-wise} \\ for Activation}} & QuEST & 4/4 & 11.76 \\
    & & Ours($h$=2) & 4/4 & - \\
    \cmidrule(lr){2-5}
    & \multirow{2}{*}{\makecell{ \textbf{Layer-wise} \\ for Activation}} & QuEST & 4/4 & 13.03 \\
    & & Ours($h$=2) & 4/4 & \textbf{8.81} \\
    \midrule
    \bottomrule
    \end{tabular}
    \caption{Comparison with EfficientDM and QuEST under specific settings. ‘Prec. (W/A)’ denotes the quantization bit-width. $h$ denotes the size of LoRA Hub.}
    \label{Standard}
\end{table}
\vspace{-10pt}

\section{Additional Visualization Results}
\label{Additional Visualization Results}

\begin{figure*}[b]
    \centering
    \vspace{-100pt}
    \includegraphics[width=1\linewidth]{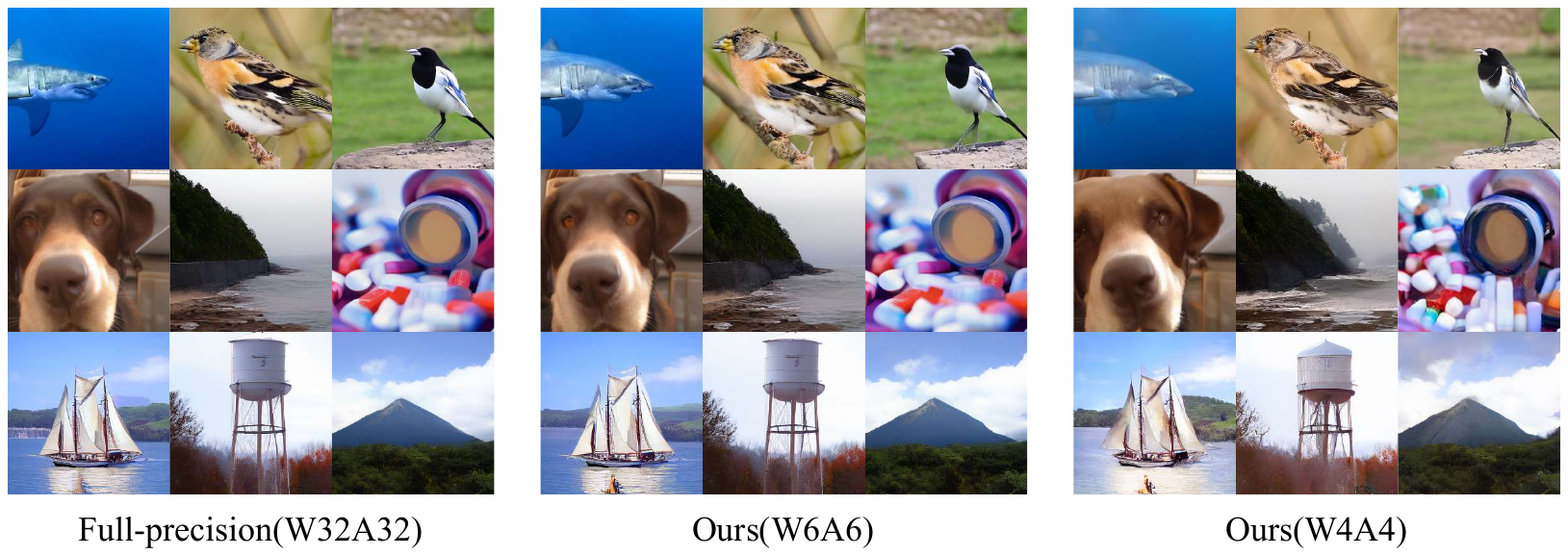}
    \vspace{-160pt} 
    \caption{Visualization of random samples from quantized LDM-4 on ImageNet. The size of LoRA Hub is 2.}
    \label{fig:ImageNet}
\end{figure*}

\begin{figure*}
    \centering
    \vspace{-20pt}
    \includegraphics[width=1\linewidth]{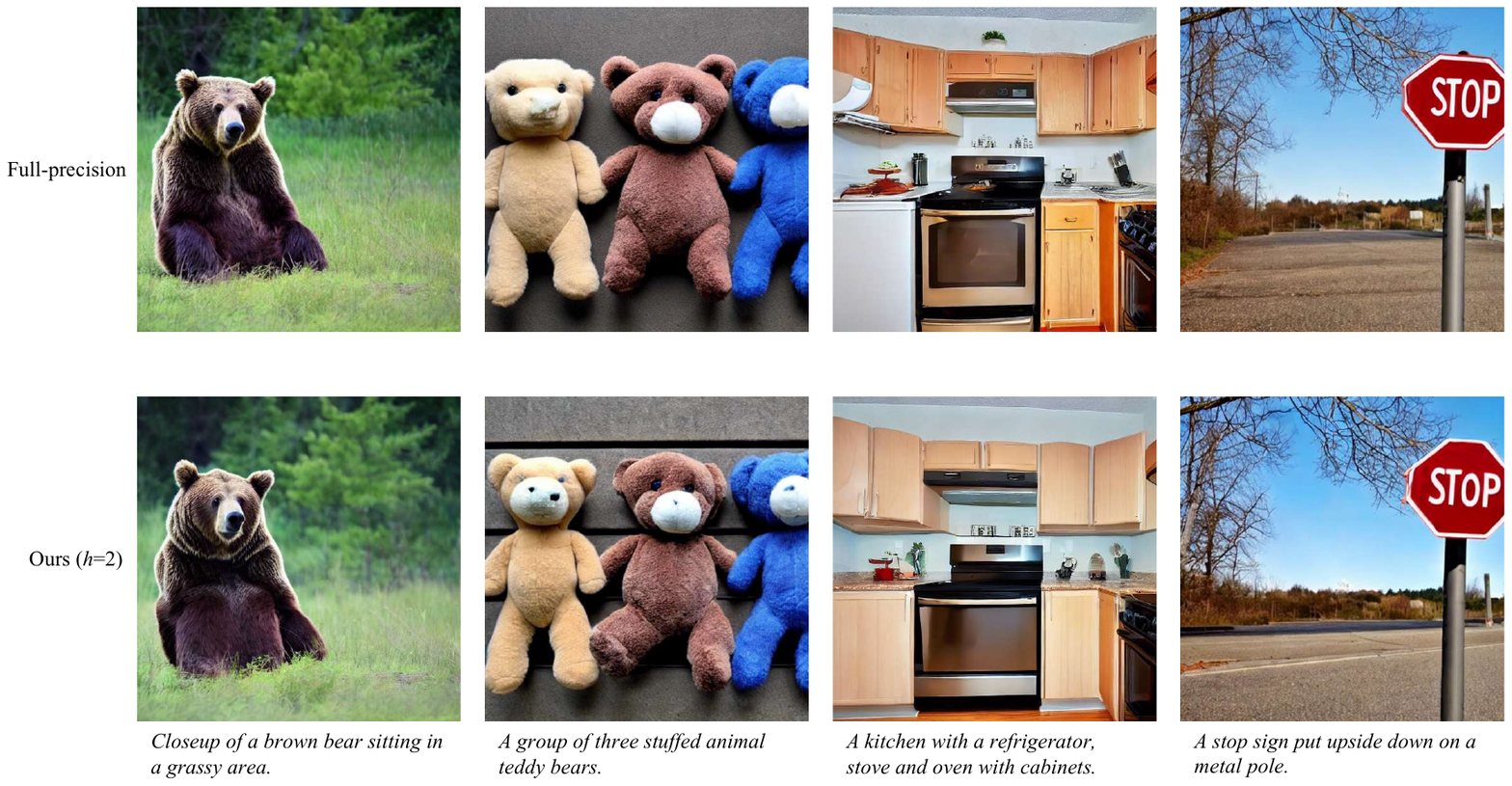}
    \vspace{-70pt} 
    \caption{Comparison of text-to-image outputs from 6-bit quantized and full-precision Stable Diffusion models. $h$ denotes the size of LoRA Hub.}
    \label{fig:SD}
\end{figure*}
